\documentclass[10pt,twocolumn,letterpaper]{article}

\usepackage[pagenumbers]{cvpr} % To force page numbers, e.g. for an arXiv version

\usepackage{algorithmicx}
\usepackage{algorithm}
\usepackage{algpseudocode}
\usepackage{graphicx}
\usepackage{amsmath}
\usepackage{amssymb}
\usepackage{booktabs}
\usepackage{bm}
\usepackage{dsfont}
\usepackage{multirow}
\usepackage{makecell}
\usepackage{gensymb}
\usepackage{xcolor}
\usepackage{scrextend}
\makeatletter
\@namedef{ver@everyshi.sty}{}
\makeatother
\usepackage{tikz}

\usepackage[pagebackref,breaklinks,colorlinks]{hyperref}

\usepackage[capitalize]{cleveref}
\crefname{section}{Sec.}{Secs.}
\Crefname{section}{Section}{Sections}
\Crefname{table}{Table}{Tables}
\crefname{table}{Tab.}{Tabs.}

\definecolor{babyblueeyes}{rgb}{0.63, 0.79, 0.95}
\definecolor{bananamania}{rgb}{0.98, 0.91, 0.71}
\definecolor{bittersweet}{rgb}{1.0, 0.44, 0.37}
\definecolor{gold}{RGB}{221,196,65}
\definecolor{silver}{RGB}{215,215,215}
\definecolor{bronze}{RGB}{126,66,5}
\newcommand{\markgold}{\tikz\draw[gold,fill=gold] (0,0) circle (.6ex);}
\newcommand{\marksilver}{\tikz\draw[silver,fill=silver] (0,0) circle (.6ex);}
\newcommand{\markbronze}{\tikz\draw[bronze,fill=bronze] (0,0) circle (.6ex);}

\definecolor{superpixel1}{RGB}{94,249,97}
\definecolor{superpixel2}{RGB}{166,198,28}
\definecolor{superpixel3}{RGB}{62,52,11}
\definecolor{superpixel4}{RGB}{169,53,140}
\definecolor{superpixel5}{RGB}{33,130,68}
\newcommand{\spa}{\tikz\draw[superpixel1,fill=superpixel1] (0,0) rectangle ++(6pt,6pt);}
\newcommand{\spb}{\tikz\draw[superpixel2,fill=superpixel2] (0,0) rectangle ++(6pt,6pt);}
\newcommand{\spc}{\tikz\draw[superpixel3,fill=superpixel3] (0,0) rectangle ++(6pt,6pt);}
\newcommand{\spd}{\tikz\draw[superpixel4,fill=superpixel4] (0,0) rectangle ++(6pt,6pt);}
\newcommand{\spe}{\tikz\draw[superpixel5,fill=superpixel5] (0,0) rectangle ++(6pt,6pt);}

\def\cvprPaperID{970} % *** Enter the CVPR Paper ID here
\def\confName{CVPR}
\def\confYear{2023}

\begin{document}

\title{HelixSurf: A Robust and Efficient Neural Implicit Surface Learning of Indoor Scenes with Iterative Intertwined Regularization}

\author{
    Zhihao Liang\textsuperscript{*},
    Zhangjin Huang\textsuperscript{*},
    Changxing Ding and
    Kui Jia\textsuperscript{\textdagger} \\
    South China University of Technology \\
    {\tt\small \{eezhihaoliang, eehuangzhangjin\}@mail.scut.edu.cn},
    {\tt\small \{chxding, kuijia\}@scut.edu.cn},
}
\maketitle

\let\thefootnote\relax\footnote{* indicates equal contribution.}

\let\thefootnote\relax\footnote{\textsuperscript{\textdagger}Correspondence to Kui Jia $<$kuijia@scut.edu.cn$>$.}

%%%%%%%%% ABSTRACT
\begin{abstract}
Recovery of an underlying scene geometry from multi-view images stands as a long-time challenge in computer vision research. The recent promise leverages neural implicit surface learning and differentiable volume rendering, and achieves both the recovery of scene geometry and synthesis of novel views, where deep priors of neural models are used as an inductive smoothness bias. While promising for object-level surfaces, these methods suffer when coping with complex scene surfaces. 
In the meanwhile, traditional multi-view stereo can recover the geometry of scenes with rich textures,  by globally optimizing the local, pixel-wise correspondences across multiple views.
We are thus motivated to make use of the complementary benefits from the two strategies, and propose a method termed Helix-shaped neural implicit Surface learning or HelixSurf; HelixSurf uses the intermediate prediction from one strategy as the guidance to regularize the learning of the other one, and conducts such intertwined regularization iteratively during the learning process.
We also propose an efficient scheme for differentiable volume rendering in HelixSurf. 
Experiments on surface reconstruction of indoor scenes show that our method compares favorably with existing methods and is orders of magnitude faster, even when some of existing methods are assisted with auxiliary training data. The source code is available at \href{https://github.com/Gorilla-Lab-SCUT/HelixSurf}{https://github.com/Gorilla-Lab-SCUT/HelixSurf}.
 
\end{abstract}

\begin{figure}[ht]
    \centering
    \begin{minipage}[t]{\columnwidth}
	\centering
	\includegraphics[width=1\textwidth]{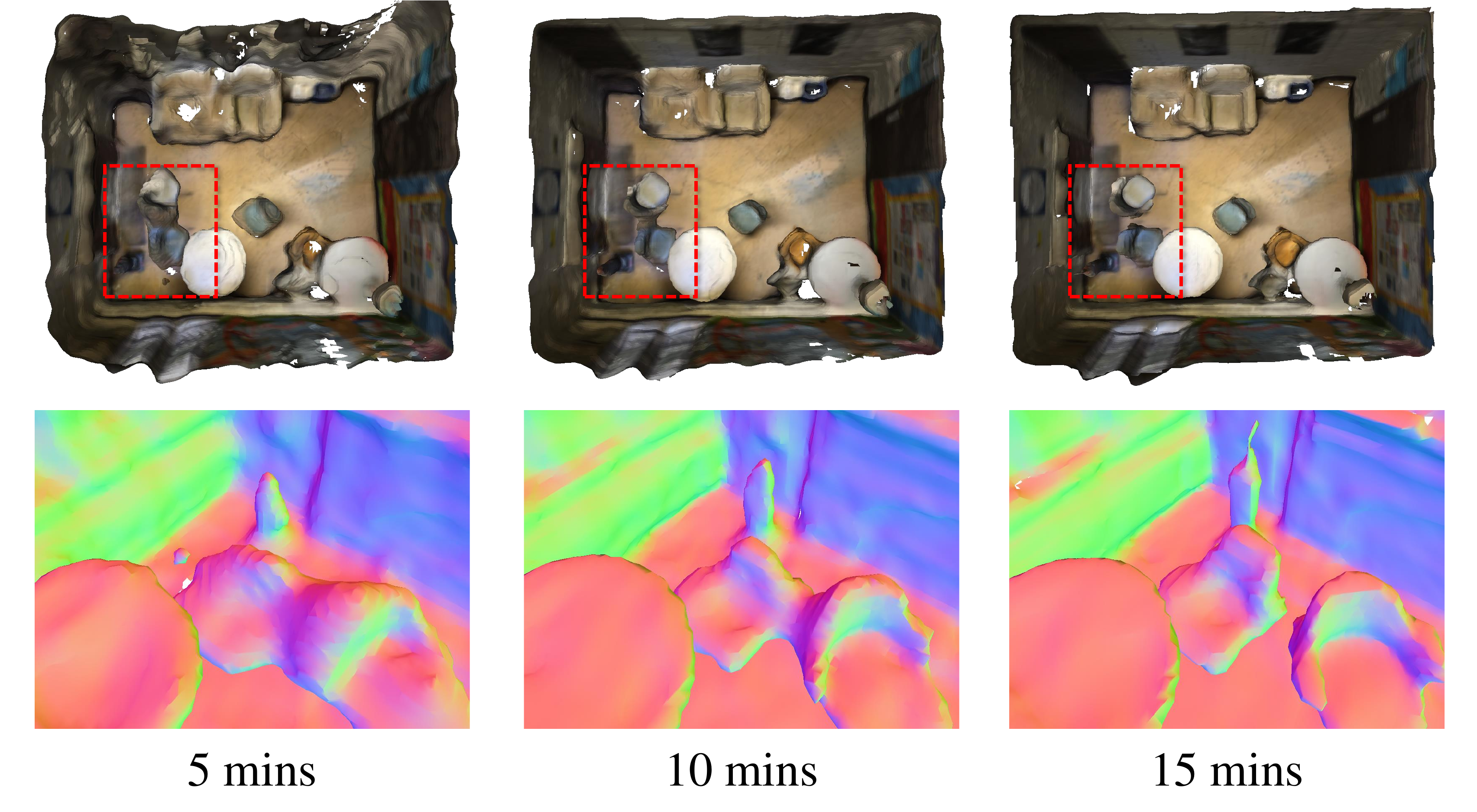}
	\subcaption{Results of HelixSurf on an example scene from ScanNet at three training checkpoints, where we use color codes to visualize surface normals.}
	\label{fig:head1}
    \end{minipage}
    \\
    \begin{minipage}[t]{\columnwidth}
	\centering
	\includegraphics[width=1\textwidth]{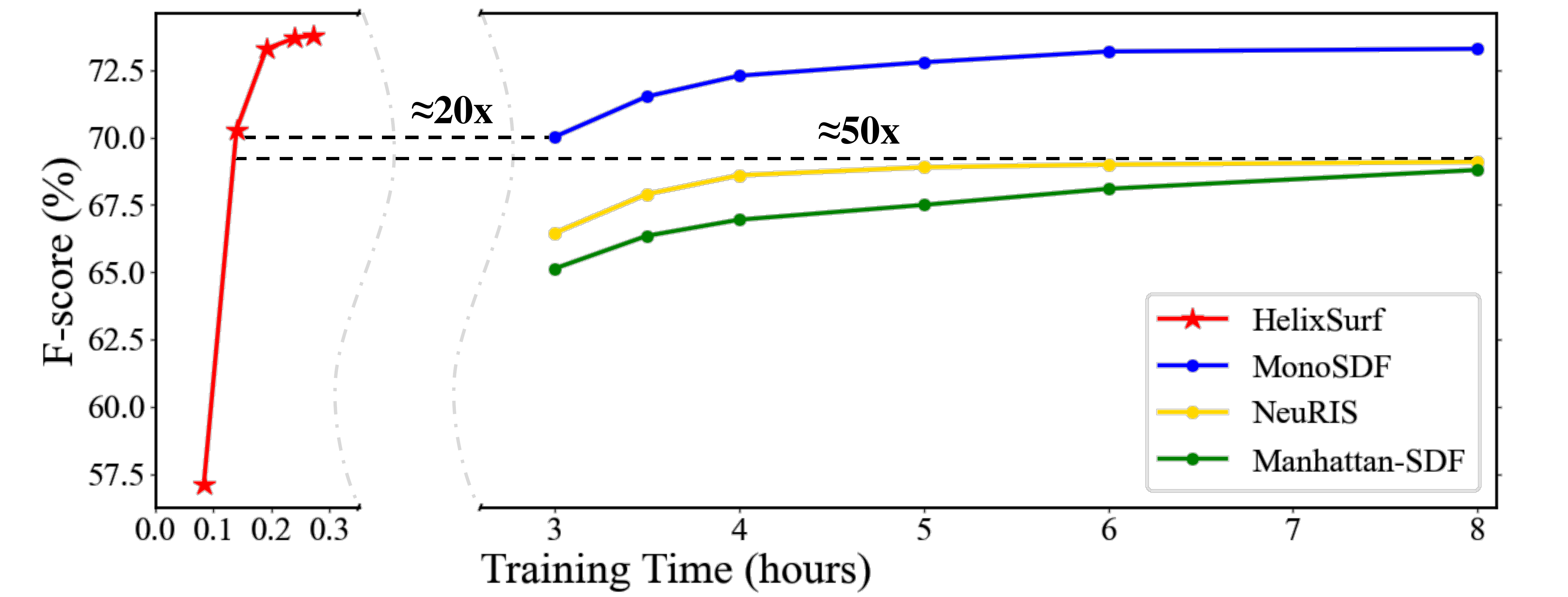}
	\subcaption{Training curves of different methods on ScanNet. The empirical training time of each method is measured on a machine with a single NVIDIA RTX 3090 GPU.}
	\label{fig:head2}
    \end{minipage}
    \caption{Efficacy and efficiency of our proposed HelixSurf. }
	\label{fig:head}
\end{figure}

%%%%%%%%% BODY TEXT
\section{Introduction}
\label{sec:intro}

Surface reconstruction of a scene from a set of observed multi-view images stands as a long-term challenge in computer vision research. A rich literature \cite{mvg_book,furukawa2015multi,chen20153survey} exists to address the challenge, including different paradigms of methods from stereo matching to volumetric fusion. Among them, the representative methods of multi-view stereo (MVS) \cite{xu2020planar, galliani2015massively,schonberger2016pixelwise,zheng2014patchmatch} first recover the properties (e.g, depth and/or normal) of discrete surface points, by globally optimizing the local, pixel-wise correspondences across the multi-view images, where photometric and geometric consistencies across views are used as the optimization cues, and a continuous fitting method (e.g., Poisson reconstruction \cite{kazhdan2006poisson, kazhdan2013screened}) is then applied to recover a complete surface. MVS methods usually make a reliable recovery only on surface areas with rich textures. 

More recently, differentiable volume rendering is proposed that connects the observed multi-view images with neural modeling of the implicit surface and radiance field \cite{mildenhall2020nerf, wang2021neus, yariv2021volume}. They show a surprisingly good promise for recovery of object-level surfaces, especially when the object masks are available in the observed images \cite{yariv2020multiview, liu2020dist}; indeed, these methods favor a continuous, closed surface given that a single deep network is used to model the scene space, whose deep prior induces a smoothness bias for surface recovery \cite{wang2021neus, yariv2021volume}. For complex scene surfaces, however, the induced smoothness bias is less capable to regularize the learning and recover the scene surface with fine geometry details \cite{zhang2020nerf++, peng2020convolutional}.  
 
To overcome the limitation, we observe that the strategies from the two paradigms of MVS and neural implicit learning are different but potentially complementary to the task. We are thus motivated to make use of the complementary benefits with an integrated solution. In this work, we achieve the goal technically by using the intermediate prediction from one strategy as the guidance to regularize the learning/optimization of the other one, and conducting such \emph{intertwined regularization iteratively} during the process. Considering that the iterative intertwined regularization makes the optimization curve as a shape of double helix, we term our method as \emph{Helix-shaped neural implicit Surface learning or HelixSurf}. Given that MVS predictions are less reliable for textureless surface areas, we regularize the learning on such areas in HelixSurf by leveraging the homogeneity inside individual superpixels of observed images. We also improve the efficiency of differentiable volume rendering in HelixSurf, by maintaining dynamic occupancy grids that can adaptively guide the point sampling along rays; our scheme improves the learning efficiency with orders of magnitude when compared with existing neural implicit surface learning methods, even with the inclusion of MVS inference time. An illustration of the proposed HelixSurf is given in \cref{fig:pipeline}. Experiments on the benchmark datasets of ScanNet\cite{dai2017scannet} and Tanks and Temples\cite{knapitsch2017tanks} show that our method compares favorably with existing methods, and is orders of magnitude faster. We note that a few recent methods \cite{wang2022neuris, Yu2022MonoSDF} use geometric cues provided by models pre-trained on auxiliary data to regularize the neural implicit surface learning; compared with them, our method achieves better results as well. Our technical contributions are summarized as follows. 

\begin{itemize}
\item We present a novel method of \emph{HelixSurf} for reconstruction of indoor scene surface from multi-view images. HelixSurf enjoys the complementary benefits of the traditional MVS and the recent neural implicit surface learning, by regularizing the learning/optimization of one strategy iteratively using the intermediate prediction from the other;
\item MVS methods make less reliable predictions on textureless surface areas. We further devise a scheme that regularizes the learning on such areas by leveraging the region-wise homogeneity organized by superpixels in each observed image;  
\item To improve the efficiency of differentiable volume rendering in HelixSurf, we adopt a scheme that can adaptively guide the point sampling along rays by maintaining dynamic occupancy grids in the 3D scene space; our scheme improves the efficiency with orders of magnitude when compared with existing neural implicit surface learning methods.
\end{itemize}

%------------------------------------------------------------------------
\section{Related Works}
\label{sec:relworks}

%-------------------------------------------------------------------------
\subsection{PatchMatch based Multi-view Stereo}
3D reconstruction from posed multi-view images is a fundamental but challenging task in computer vision.
Among all the techniques in the literature, PatchMatch based Multi-view Stereo (PM-MVS) is traditionally the most explored one \cite{mvg_book, furukawa2015multi}.
PM-MVS methods \cite{schonberger2016structure, schonberger2016pixelwise, zheng2014patchmatch, shen2013accurate,galliani2015massively, romanoni2019tapa, xu2020planar} represent the geometric with depth and/or normal maps.
They estimate depth and/or normal of each pixel by exploiting inter-image photometric and geometric consistency and then fuse all the depth maps into a global point cloud with filtering operations, which can be subsequently processed using meshing algorithms \cite{labatut2007efficient, kazhdan2006poisson}, \eg Screened Poisson surface reconstruction \cite{kazhdan2013screened}, to recover complete surface.
These traditional methods have achieved great success on various occasions and can produce plausible geometry of textured surfaces, but there exist artifacts and missing parts in the areas without rich textures.
Indeed, their optimization highly relies on the photometric measure to discriminate which random estimate is the best guess. 
In the case of indoor scenes with textureless areas \cite{xu2020planar,romanoni2019tapa}, the inherent homogeneity inactivates the photometric measure and consequently poses difficulties to the accurate depth estimation.
With the development of deep learning, learning-based MVS methods \cite{yao2018mvsnet, yao2019recurrent, im2018dpsnet, xu2020pvsnet, wang2021patchmatchnet} demonstrate promising performance in recent years.
However, they crucially rely on ground-truth 3D data for supervision, which hinders their practical application.

%-------------------------------------------------------------------------
\subsection{Neural Implicit Surface}
In contrast to classic explicit representation, recent works \cite{park2019deepsdf,mescheder2019occupancy,chibane2020neural} implicitly represent surfaces via learning neural networks, which models continuous surface with Multi-Layer Perceptron (MLP) and makes it more feasible and efficient to represent complex geometries with arbitrary typologies.
For the task of multi-view reconstruction, the 3D geometry is represented by a neural network that outputs either a signed/unsigned distance field or an occupancy field. 
Some works \cite{niemeyer2020differentiable, yariv2020multiview, liu2020dist} utilize surface rendering to enable the reconstruction of 3D shapes from 2D images, but they always rely on extra object masks.
Inspired by the success of NeRF \cite{mildenhall2020nerf}, recent works \cite{wang2021neus, oechsle2021unisurf, yariv2021volume} attach differentiable volume rendering techniques to reconstruction, which eliminates the need of mask and achieves impressive reconstruction.
And follow-up works \cite{fu2022geo, darmon2022improving, wang2022hfneus} further improve the geometry quality with fine-grained surface details.
Although these methods show better accuracy and completeness compared with the traditional MVS methods, they still suffer from the induced smoothness bias of deep network \cite{peng2020convolutional, zhang2020nerf++}, which discourages them to regularize the learning and recover fine details in scene reconstruction.
Most recent works \cite{guo2022neural, wang2022neuris, Yu2022MonoSDF} try to get rid of this dilemma by incorporating geometric cues provided by models pre-trained on auxiliary data.
Our HelixSurf integrates traditional PM-MVS and neural implicit learning surface in complementary mechanisms and achieves better results than these methods.

%-------------------------------------------------------------------------

\begin{figure*}
    \centering
    \includegraphics[width=1.0\textwidth]{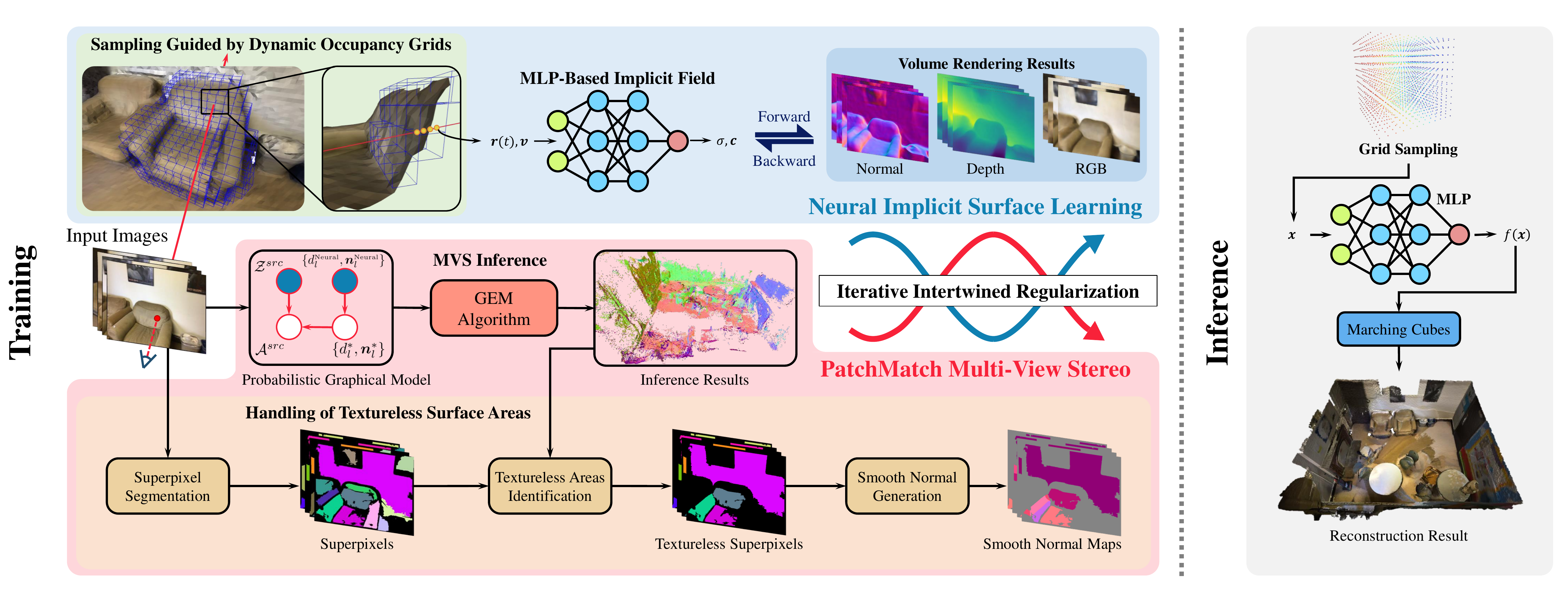}
    \caption{\textbf{Overview of HelixSurf: Helix-shaped neural implicit Surface learning}. HelixSurf integrates the neural implicit surface learning (\cf Section \ref{subsec:reg_neural_from_MVS}) and PatchMatch based MVS (\cf Section \ref{subsec:reg_MVS_from_neural}) in a robust and efficient manner. We optimize HelixSurf with an iterative intertwined regularization, which uses the intermediate prediction from one strategy as guidance to regularize the learning/optimization of the other one; given that MVS predictions are less reliable for textureless surface areas, we additionally devise a scheme that regularizes the learning on such areas by leveraging the homogeneity per superpixel in observed multi-view images (\cf Section \ref{subsec:handling_textureless_areas}).
    We also propose a scheme for point sampling along rays (\cf Section \ref{subsec:occupancy_grids}), which significantly improves the efficiency. At the inference stage of HelixSurf, we conduct grid sampling to query the learned SDF values at sampled points and run Marching Cubes to get the reconstruction results.
    }
\label{fig:pipeline}
\end{figure*}

\section{Preliminary}
\label{sec:preliminary}

In this section, we give technical background and math notations that are necessary for presentation of our proposed method in subsequent sections.

\vspace{0.1cm}
\noindent{\textbf{Neural Implicit Surface Representation}}
Among the choices of neural implicit surface representation \cite{park2019deepsdf,mescheder2019occupancy,chibane2020neural}, we adopt DeepSDF \cite{park2019deepsdf} that learns to encode a continuous surface as the zero-level set of a signed distance field (SDF) $f: \mathbb{R}^3 \rightarrow \mathbb{R}$, which is typically parameterized as an MLP; for any point $\bm{x}\in\mathbb{R}^3$ in the 3D space, $| f(\bm{x}) |$ assigns its distance to the surface $\mathcal{S} = \left\{\bm{x} \in \mathbb{R}^3 | f(\bm{x}) = 0\right\}$; by convention, we have $f(\bm{x}) < 0$ for points inside the surface and $f(\bm{x}) > 0$ for those outside.

\vspace{0.1cm}
\noindent{\textbf{SDF-induced Volume Rendering}}
Differentiable volume rendering is used in NeRF \cite{mildenhall2020nerf} for synthesis of novel views.
Denote a ray emanating from a viewing camera as $\bm{r}(t) = \bm{o} + t\bm{v}$, $t\geq 0$, where $\bm{o}\in\mathbb{R}^3$ is the camera center and $\bm{v}\in\mathbb{R}^3, \Vert \bm{v} \Vert = 1$ denotes the unit vector of viewing direction.
NeRF models a continuous scene space as a neural radiance field $\bm{F}: \mathbb{R}^3 \times \mathbb{R}^3 \rightarrow \mathbb{R}_{+} \times \mathbb{R}^3$, which for any space point $\bm{x}$ and direction $\bm{v}$, assigns $ \bm{F}(\bm{x}, \bm{v}) = (\sigma, \bm{c})$, where $\sigma \in \mathbb{R}_{+}$ represents the volume density at the location $\bm{x}$, and $\bm{c} \in \mathbb{R}^3$ is the view-dependent color from $\bm{x}$ along the ray $- \bm{r}$ towards $\bm{o}$.
Assume $N$ points are sampled along $\bm{r}$; the color accumulated along the ray $\bm{r}$ can be approximated, using the quadrature rule\cite{max1995optical}, as
\begin{equation}
    \label{eq:render_color}
    \bm{C}(\bm{r}) = \sum^N_{i = 1} T_i \alpha_i \bm{c}(\bm{r}(t_i), \bm{v}),\quad
    T_i = \prod^{i - 1}_{j = 1}(1 - \alpha_j) ,
\end{equation}
where $\alpha_i = 1 - \exp(-\int^{t_{i + 1}}_{t_i}\sigma(\bm{r}(t)) dt)$ denotes the opacity of a segment. While the volume density $\sigma: \mathbb{R}^3 \rightarrow \mathbb{R}_{+}$ is learned as a direct output of the MLP based radiance field function $\bm{F}$ in \cite{mildenhall2020nerf}, it is shown in VolSDF \cite{yariv2021volume} and NeuS \cite{wang2021neus} that $\sigma$ can be modeled as a transformed function of the implicit SDF function $f$, enabling better recovery of the underlying geometry. In this work, we follow \cite{wang2021neus} to model $\sigma$ as an SDF-induced volume density. With such an SDF-induced, differentiable volume rendering, the geometry $f$ and color $\bm{c}$ can be learned by minimizing the difference between rendering results and multiple views of input images. Note that analogous to (\ref{eq:render_color}), the depth $d$ of the surface from the camera center $\bm{o}$ can be approximated along the ray $\bm{r}$ as well, giving rise to
\begin{equation}
\label{eq:render_depthnorm}
    d(\bm{r}) = \sum^N_{i = 1} T_i \alpha_i t_i,
    \quad
    \bm{n}(\bm{r}) = \nabla f(\bm{o} + d(\bm{r}) \bm{v}) ,
\end{equation}
where $\bm{n}(\bm{r}) \in \mathbb{R}^3$ denotes the surface normal at the intersection point and $\nabla f(\bm{x})$ is the gradient of SDF at $\bm{x}$.

\vspace{0.1cm}
\noindent{\textbf{Multi-View Stereo with PatchMatch}}
Assume that a reference image $\bm{I}^\text{ref}$ and a set of source images $\mathcal{I}^\text{src} = \{\bm{I}^m | m = 1\ldots M\}$ capture a common scene; we write collectively as $\mathcal{I} = \{\bm{I}^\text{ref}, \mathcal{I}^\text{src} \}$.
PatchMatch based multi-view stereo (PM-MVS) methods\cite{zheng2014patchmatch, schonberger2016pixelwise, galliani2015massively, xu2020planar} aim to recover the scene geometry by predicting the depth $d_l \in \mathbb{R}^+$ and normal $\bm{n}_l \in \mathbb{R}^3, \Vert \bm{n}_l \Vert = 1$ for each pixel in $\bm{I}^\text{ref}$, which is indexed by $l$ with $l \in \{1, \dots, L\}$.
Considering that any $l^{th}$ pixel in $\bm{I}^\text{ref}$ may not be visible in all images in $\mathcal{I}^\text{src}$, the methods then predict an occlusion indicator $\mathcal{Z}^{src} = \{Z^m_l | l = 1, \ldots, L, m = 1, \ldots, M\}$ for $\bm{I}^\text{ref}$. Optimization of $\{d_l\}_{l=1}^L$, $\{\bm{n}_l\}_{l=1}^L$, and $\mathcal{Z}^{src}$ is based on enforcing photometric and geometric consistencies between corresponding patches in $\bm{I}^\text{ref}$ and $\mathcal{I}^\text{src}$; this is mathematically formulated as a probabilistic graphical model and is solved via generalized expectation-maximization (GEM) algorithm \cite{schonberger2016pixelwise, galliani2015massively}, where PatchMatch \cite{barnes2009patchmatch,bleyer2011patchmatch} is used to efficiently establish pixel-wise correspondences across multi-view images.
More specifically, let $\mathcal{A}^{src} = \{ \bm{A}^m_l | l = 1, \ldots, L, m = 1, \ldots, M\}$ denote the set of homography-warped patches from source images \cite{shen2013accurate}, the PatchMatch based methods optimize $d_l$ and $\bm{n}_l$ for a pixel in the reference image as
\begin{equation}
\begin{aligned}
\label{eq:mvs_em}
    \{d_l^{*}, \bm{n}_l^{*}\}
    &= \mathop{\arg\max}P( d_l, \bm{n}_l | \mathcal{A}^{src}, \mathcal{Z}^{src})  \\
    &\propto \mathop{\arg\max}P( \mathcal{A}^{src} | d_l, \bm{n}_l, \mathcal{Z}^{src})P(d_l, \bm{n}_l)  \\
    &= \mathop{\arg\min} \sum_{m=1}^{M} P_l(m) \xi^m_l(d_l, \bm{n}_l) \\
    \textrm{with} \ \  \xi^m_l =& \ 1 - \rho^m_l (d_l, \bm{n}_l) + \eta\min(\varphi^m_l (d_l, \bm{n}_l), \varphi_\text{max}) ,
\end{aligned}
\end{equation}
where $\rho^m_l(d_l, \bm{n}_l)$ denotes the color similarity between the reference patch $\bm{A}^\text{ref}_l$ and source patch $\bm{A}^{m}_l$ based on normalized cross-correlation, which is a function of $d_l$ and $\bm{n}_l$, and $\varphi^m_l(d_l, \bm{n}_l)$ is the forward-backward reprojection error to evaluate the geometric consistency incurred by the predicted $d_l$ and $\bm{n}_l$, which is capped by a pre-defined $\varphi_\text{max}$; the probability $P_l(m)$ serves for view selection that assigns different weights to the $M$ source images.
Indeed, source images with small values of $P_l(m)$ are less informative; hence Monte-Carlo view sampling is used in \cite{zheng2014patchmatch} to draw samples according to $P_l(m)$. Assume that the selected views form a subset $S \subset \{1\ldots M\}$, the problem (\ref{eq:mvs_em}) can be simplified as
\begin{equation}
    \{d_l^{*}, \bm{n}_l^{*}\} = \mathop{\arg\min}  \frac{1}{|S|} \sum_{m\in S} \xi^m_l(d_l, \bm{n}_l) .
\end{equation}

%-------------------------------------------------------------------------

\section{HelixSurf for Intertwined Regularization of Neural Implicit Surface Learning}
\label{sec:methods}

Given a set of calibrated RGB images $\{\bm{I}_m\}_{m=1}^M$ of an indoor scene captured from multiple views, the task is to reconstruct the scene geometry with fine details.
Under the framework of neural differentiable volume rendering, the task translates as learning an MLP based radiance field function $\bm{F}$ that connects the underlying scene geometry with the image observations $\{\bm{I}_m\}_{m=1}^M$; with the use of an SDF-induced volume density $\sigma(f)$, the scene surface can be reconstructed by extracting the zero-level set of the learned SDF $f$.
As stated in Section \ref{sec:intro}, although the supervision from $\{\bm{I}_m\}_{m=1}^M$ is conducted in a pixel-wise, independent manner, the MLP based function $f$ has \emph{deep priors} that induce the function learning biased towards encoding continuous and piece-wise, smooth surface \cite{wang2021neus,yariv2021volume}; indeed, assuming a successful learning of a ReLU-based MLP $f$, its zero-level set can be \emph{exactly} recovered as a continuous polygon mesh \cite{lei2020analytic}. In the meanwhile, PatchMatch based MVS methods couple the predictions of $\{ d_l, \bm{n}_l \}$ for individual pixels in a probabilistic framework, and conduct the optimization \emph{globally} such that the predicted $\{ d_l, \bm{n}_l \}$ achieves an overall best consistencies of photometry and geometry across $\{\bm{I}_m\}_{m=1}^M$; after obtaining $\{ d_l, \bm{n}_l \}$, a continuous, watertight surface can be fitted using Poisson reconstruction \cite{kazhdan2006poisson, kazhdan2013screened}.
The above two strategies reconstruct the surface using different but potentially complementary mechanisms. We are thus motivated to propose an integrated solution that can take both advantages of them. In this work, we achieve the goal technically by using the intermediate prediction from one strategy as the guidance to regularize the learning of the other one, and conducting such \emph{intertwined regularization iteratively} during the learning process. Considering that the iterative intertwined regularization makes the optimization curve as a shape of double helix, we term our method as \emph{Helix-shaped neural implicit Surface learning or HelixSurf}.
Details of HelixSurf are presented as follows. An illustration is given in \cref{fig:pipeline}.

\subsection{Regularization of Neural Implicit Surface Learning from MVS predictions}
\label{subsec:reg_neural_from_MVS}

Given the set of multi-view images $\mathcal{I} = \{\bm{I}^\text{ref}, \mathcal{I}^\text{src} \}$, neural implicit surface learning via differentiable volume rendering samples rays in the 3D space; for any sampled ray $\bm{r}(t) = \bm{o} + t\bm{v}$, $t \geq 0$, in a viewing direction $\bm{v}$, assume that it emanates from the camera center $\bm{o}$ and passes through a pixel $\bm{a} \in \mathbb{R}^3$ in an image $\bm{I}$ in $\mathcal{I}$. Let $\bm{F}$ be the SDF-induced neural radiance field that models the scene geometry via the SDF function $f$; we can then write as $\bm{F}(\bm{r}(t), \bm{v}; f) = (\sigma(f(\bm{r}(t))), \bm{c}(\bm{r}(t), \bm{v}))$ for any point $t$ along $\bm{r}(t)$. According to (\ref{eq:render_color}) of approximated volume rendering, the color $\bm{C}(\bm{r})$ accumulated along the ray $\bm{r}$ can be computed, given $\{ \sigma(f(\bm{r}(t_i))), \bm{c}(\bm{r}(t_i), \bm{v}) \}_{i=1}^N$ at $N$ sampled points;
the following loss defines the color based image supervision from ray $\bm{r}$ for learning $\bm{F}$ (i.e., learning the MLPs $f$ and $\bm{c}$, see Section \ref{sec:preliminary} for the details):
\begin{equation}
\mathcal{L}_\text{\tiny Neural}(\bm{r}; f, \bm{c}) =  \texttt{SmoothL1}(\bm{C}(\bm{r}; f, \bm{c}), \bm{a}(\bm{r})) .
\end{equation}
We can also compute the depth $d(\bm{r}; f)$ and surface normal $\bm{n}(\bm{r}; f)$  according to (\ref{eq:render_depthnorm}).

Section \ref{sec:preliminary} suggests that given $\mathcal{I}$, PatchMatch based MVS methods can predict pairs of depth and surface normal for pixels in the observed reference image. Such methods usually produce a sparse set of predictions on texture-rich surface areas \cite{schonberger2016pixelwise,Xu2019ACMM}. Without loss of generality, assume that $\{d_{\bm{a}}^{\text{\tiny MVS}}, \bm{n}_{\bm{a}}^{\text{\tiny MVS}}\}$ are the MVS prediction for the pixel $\bm{a}$ in the image $\bm{I}$. We use $\{d_{\bm{a}}^{\text{\tiny MVS}}, \bm{n}_{\bm{a}}^{\text{\tiny MVS}} \}$ to regularize the learning of $f$ in the current iteration, based on the following loss
\begin{equation}
\begin{aligned}
    \mathcal{L}_\text{\tiny MVSRegu}(\bm{r}; f) &=  w(\bm{r}) \left(\bigl\lvert d(\bm{r}; f) - d_{\bm{a}}^{\text{\tiny MVS}} \bigr\rvert + \bigl\lvert \bm{n}(\bm{r}; f) - \bm{n}_{\bm{a}}^{\text{\tiny MVS}} \bigr\rvert \right) , \\
    \textrm{with} \ \  w(\bm{r}) &= \mathds{1}_{\text{\tiny MVSRegu}}(\bm{r}) \cdot (1 - \bigl\lvert \bm{C}(\bm{r}) - \bm{a}(\bm{r}) \bigr\rvert)
\end{aligned}
\end{equation}
where $\mathds{1}_{\text{\tiny MVSRegu}}(\bm{r})$ is an indicator to cope with the case when $\{d_{\bm{a}}^{\text{\tiny MVS}}, \bm{n}_{\bm{a}}^{\text{\tiny MVS}}\}$ are not predicted by MVS for the pixel $\bm{a}$.

\begin{figure}[htbp]
    \centering
    \includegraphics[width=0.47\textwidth]{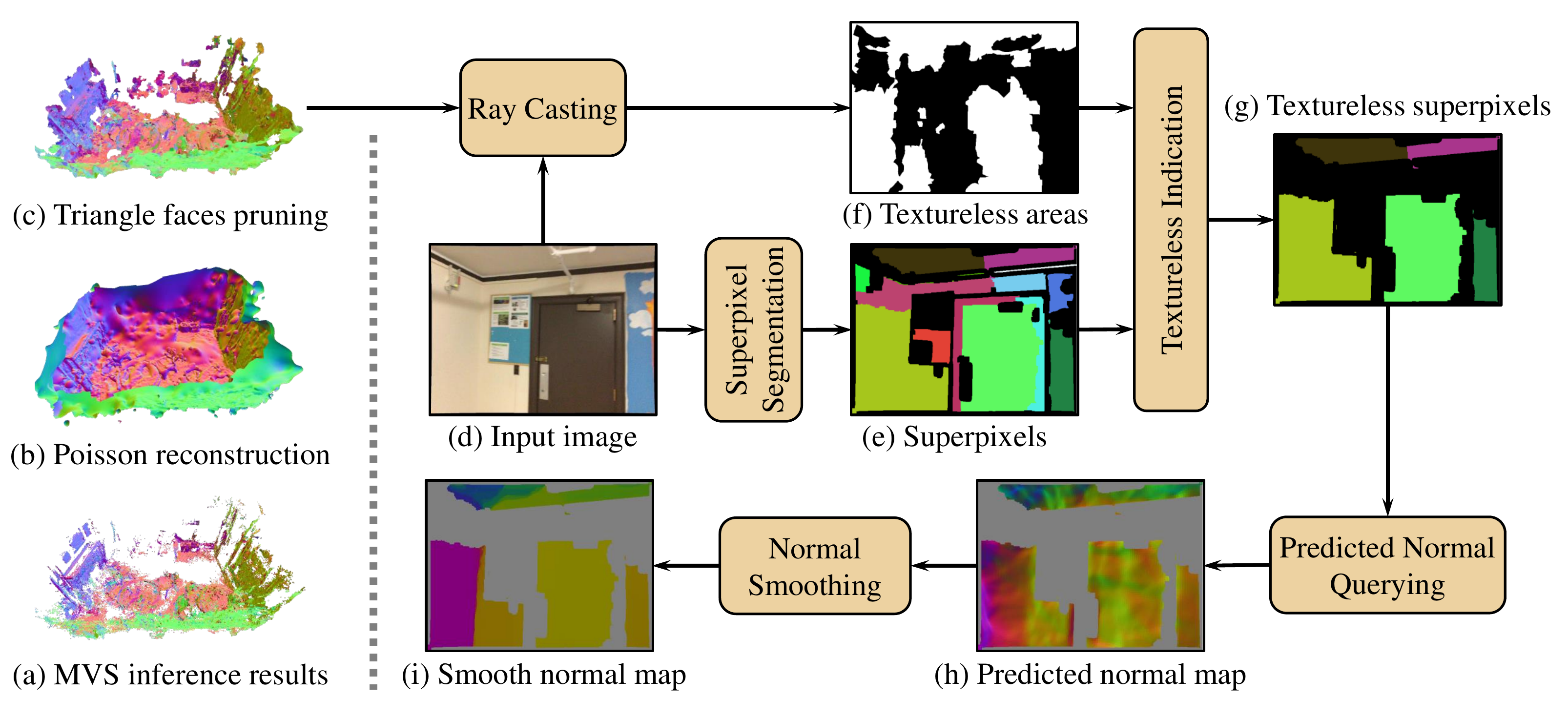}
    \caption{Illustration of handling textureless surface areas.
    (a): the inference results of PM-MVS,
    (b): watertight surface mesh $\mathcal{M^\text{\tiny MVS}}$ reconstructed by Poisson reconstruction from (a),
    (c): surface $\mathcal{M}_{-}^\text{\tiny MVS}$ obtained by pruning textureless triangle faces from (b),
    (d): an input image,
    (e): superpixels extracted by the graph-based segmentation algorithm \cite{felzenszwalb2004efficient},
    (f): textureless areas obtained by ray casting,
    (g): superpixels covered by textureless areas,
    (h): surface normal map predicted by HelixSurf,
    (i): smooth normal map obtained by aggregating the normals in textureless superpixels.
    }
\label{fig:planar_normal}
\end{figure}

\subsubsection{Handling of Textureless Surface Areas}
\label{subsec:handling_textureless_areas}

PatchMatch based MVS methods make reliable predictions only on texture-rich surface areas. We resort to other sources to regularize the neural implicit learning for texture-less surface areas. Our motivation is based on the observation that textureless surface areas tend to be both homogeneous in color and geometrically smooth; indeed, when the surface areas are of high curvature or when they have different colors, 2D image projections of such areas would have richer textures. The projected 2D image counterparts of textureless surface areas in fact correspond to those in images that can be organized as superpixels. We thus propose to further regularize the neural implicit surface learning by leveraging the homogeneity of image superpixels.

We technically encourage the predicted normals of surface points, whose 2D projections fall in a same superpixel, to be close. For any image $\bm{I}$ in $\mathcal{I}$, we pre-compute its region partitions of superpixels using methods such as \cite{felzenszwalb2004efficient,achanta2012slic}. Let $\tilde{\bm{r}}$ be a ray passing through a pixel $\tilde{\bm{a}}$ that falls in a superpixel of $\bm{I}$; denote the superpixel as $\tilde{\bm{A}}_{\tilde{\bm{a}}}$. We know that the volume rendering in HelixSurf predicts surface normal $\bm{n}(\tilde{\bm{r}}; f)$ for the ray $\tilde{\bm{r}}$, and denote as $\{ \bm{n}'(\tilde{\bm{r}}'; f) | \tilde{\bm{a}}' \in \tilde{\bm{A}}_{\tilde{\bm{a}}} \}$ the predicted surface normals for all pixels in $\tilde{\bm{A}}_{\tilde{\bm{a}}}$. We first compute $ \bm{n}_{\tilde{\bm{a}}, \bm{I}}^{\text{\tiny Smooth}} = \sum_{i=1}^{|\tilde{\bm{A}}_{\tilde{\bm{a}}}|} \bm{n}'_i / |\tilde{\bm{A}}_{\tilde{\bm{a}}}| $, and then apply the above computation to all those images in $\mathcal{I}$ that capture the same surface point and have the corresponding pixels of $\tilde{\bm{a}}$ in $\bm{I}$. Assume we have a total of $M'$ such image, we compute $\bm{n}_{\tilde{\bm{a}}}^{\text{\tiny Smooth}} = \sum_{m=1}^{M'+1} \bm{n}_{\tilde{\bm{a}}_m, \bm{I}_m}^{\text{\tiny Smooth}} / (M' + 1)$ and enforce closeness of surface normal predictions for pixels both inside a superpixel and across multi-view images with the following loss 
\begin{equation}
    \mathcal{L}_\text{\tiny Smooth}(\tilde{\bm{r}}; f) =   \mathds{1}_\text{\tiny Smooth}(\tilde{\bm{r}}) \cdot \bigl\lvert \bm{n}(\tilde{\bm{r}}; f) - \bm{n}_{\tilde{\bm{a}}}^{\text{\tiny Smooth}} \bigr\rvert  ,
\end{equation}
where $\mathds{1}_\text{\tiny Smooth}(\tilde{\bm{r}})$ indicates whether the pixel $\tilde{\bm{a}}$ cast by the ray $\tilde{\bm{r}}$ belongs to a textureless area. In practice, we identify the textureless areas for a surface $\mathcal{S} = \left\{\bm{x} \in \mathbb{R}^3 | f(\bm{x}) = 0\right\}$ as follows. We first use MVS methods to produce a sparse set of depth and normal predictions, to which we apply Poisson reconstruction \cite{kazhdan2006poisson, kazhdan2013screened} and obtain a watertight surface mesh $\mathcal{M^\text{\tiny MVS}}$ (\cref{fig:planar_normal}(b)). We prune those triangle faces in $\mathcal{M^\text{\tiny MVS}}$ that contain no the depths and normals predicted by the MVS methods, resulting in $\mathcal{M}_{-}^\text{\tiny MVS}$. For an image $\bm{I}$, we conduct ray casting and treat the pixels whose associated rays do not hit $\mathcal{M}_{-}^\text{\tiny MVS}$ as those belonging to textureless areas (\cref{fig:planar_normal}(e)). The overall scheme is illustrated in \cref{fig:planar_normal}. Please refer to the supplementary for more details.

\subsection{Regularization of Multi-View Stereo from Neural Implicit Surface Learning}
\label{subsec:reg_MVS_from_neural}

\cref{eq:mvs_em} of MVS methods optimize the depth and normal predictions by maximizing a posterior probability and a prior of $P(d, \bm{n})$ (cf. line 2 in \cref{eq:mvs_em}). Without other constraints, $P(d, \bm{n})$ is usually set as a uniformly random distribution. In HelixSurf, it is obviously feasible to use the depth and normal learned in the current iteration of neural implicit learning as the prior. 

More specifically, given $d_l$, $\bm{n}_l$, $\mathcal{A}^{src}$, and $\mathcal{Z}^{src}$ denoted as in Section \ref{sec:preliminary}, let $d_l^{\text{\tiny Neural}}$ and $\bm{n}_l^{\text{\tiny Neural}}$ be the depth and normal learned in the current iteration of neural implicit learning for the corresponding pixel in an observed image. We can improve MVS predictions using
\begin{equation}
\begin{aligned}
\label{eq:mvs_prior}
\{d_l^{*}, \bm{n}_l^{*}\} = \arg\max P(d_l, \bm{n}_l | \mathcal{A}^{src}, \mathcal{Z}^{src}, d_l^{\text{\tiny Neural}}, \bm{n}_l^{\text{\tiny Neural}}) \\ 
\propto \arg\max P( \mathcal{A}^{src} | d_l, \bm{n}_l, \mathcal{Z}^{src}) P(d_l, \bm{n}_l | d_l^{\text{\tiny Neural}}, \bm{n}_l^{\text{\tiny Neural}}) .
\end{aligned}
\end{equation}
Qualitative results in Section \ref{subsec:ablation} show that MVS methods with priors of a uniformly random distribution tend to produce noisy results with outliers, which would impair the iterative learning in HelixSurf. Instead, the proposed (\ref{eq:mvs_prior}) gives better results.

\subsection{Improving the Efficiency by Establishing Dynamic Space Occupancies}
\label{subsec:occupancy_grids}

Differentiable volume rendering suffers from the heavy cost of point sampling along rays for accumulating pixel colors \cite{mildenhall2020nerf,wang2021neus,yariv2021volume}. While a common coarse-to-fine sampling strategy is used in these methods, it still counts as the main computation. In this work, we are inspired by Instant-NGP and propose a simple yet effective sampling scheme, which establishes dynamic occupancies in the 3D scene space and adaptively guides the point sampling along rays. \cref{fig:pipeline} gives the illustration.

More specifically, we partition the 3D scene space regularly using a set $\mathcal{G}_\text{\tiny{Occu}}$ of occupancy grids of size $64^3$, and let the occupancy of any voxel partitioned and indexed by $\{ g \subset \mathcal{G}_\text{\tiny{Occu}} \}$  be $o_g$.
During training of HelixSurf, we update $o_g$ using exponential moving average (EMA), i.e., $o_g^{\text{\tiny EMA}} \leftarrow \max(\sigma_g, \alpha (\sigma_g - o_g^{\text{\tiny EMA}}) + o_g^{\text{\tiny EMA}} )$, where $\sigma_g$ is the density at $g$ given by the inducing SDF function $f$ and $\alpha = 0.05$ is a decaying factor. We set the voxel indexed by $g$ as occupied if $o_g^{\text{\tiny EMA}} > \tau_\text{\tiny{Occu}}$, where $\tau_\text{\tiny{Occu}}$ is a pre-set threshold. Non-occupied voxels will be skipped directly when performing point sampling along each ray, thus improving the efficiency of differentiable volume rendering used in HelixSurf. More details of our scheme are given in the supplementary material.
\cref{fig:head}(b) shows that our scheme improves the training efficiency at orders of magnitude when compared with existing neural implicit surface learning methods.

\subsection{Training and Inference}
\label{subsec:training_and_inference}
At each iteration of HelixSurf training, we randomly sample pixels from the images in $\mathcal{I}$ and define the set of camera rays passing through these pixels as $\mathcal{R} \cup \tilde{\mathcal{R}}$, where $\mathcal{R}$ and $\tilde{\mathcal{R}}$ contain rays passing through texture-rich and textureless areas respectively. We optimize the following problem to learn the MLP based functions $f$ and $\bm{c}$

\begin{equation}
\begin{aligned}
\min_{f, \bm{c}} (& \sum_{\bm{r} \in \mathcal{R}} \mathcal{L}_\text{\tiny Neural}(\bm{r}; f, \bm{c}) + \sum_{\tilde{\bm{r}} \in \tilde{\mathcal{R}}} \mathcal{L}_\text{\tiny Neural}(\tilde{\bm{r}}; f, \bm{c}) +  \\ & \lambda_\text{\tiny MVSRegu} \sum_{\bm{r} \in \mathcal{R}} \mathcal{L}_\text{\tiny MVSRegu}(\bm{r}; f)  + \lambda_\text{\tiny Smooth} \sum_{\tilde{\bm{r}} \in \tilde{\mathcal{R}}} \mathcal{L}_\text{\tiny Smooth}(\tilde{\bm{r}}; f)  \\ & \quad\quad\quad \lambda_\text{\tiny Eik} \sum_{\bm{x} \in \mathbb{R}^3} \mathcal{L}_\text{Eik}(\bm{x}; f) ),
\end{aligned}
\end{equation}
where $\mathcal{L}_\text{Eik}(\bm{x}; f)$ is the Eikonal loss \cite{gropp2020implicit} that regularizes the learning of SDF $f$, and $\lambda_\text{\tiny MVSRegu}, \lambda_\text{\tiny Smooth}, \lambda_\text{\tiny Eik}$ are hyperparameters weighting different loss terms.

During inference, we apply marching cubes \cite{lorensen1987marching} algorithm to extract the underlying surface from the learned SDF $f$.

% \newpage

%-------------------------------------------------------------------------
\section{Experiments}
\label{sec:experiments}

\noindent{\textbf{Datasets}}
We conduct experiments using the benchmark dataset of ScanNet \cite{dai2017scannet} and Tanks and Temples \cite{knapitsch2017tanks}. ScanNet has 1613 indoor scenes with precise camera calibration parameters and surface reconstructions via the state-of-the-art SLAM technique \cite{dai2017bundlefusion}. Tanks and Temples has multiple large-scale indoor and outdoor scenes.
For the ScanNet, we follow ManhattanSDF\cite{guo2022neural} and select 4 scenes to conduct our experiments.
As for Tanks and Temples, we follow MonoSDF\cite{Yu2022MonoSDF} to select four large-scale indoor scenes to further investigate the extensibility of HelixSurf.

\noindent{\textbf{Implementation Details}}
We implement HelixSurf in PyTorch\cite{paszke2019pytorch} framework with CUDA extensions, and customized a PM-MVS module for HelixSurf according to COLMAP \cite{schonberger2016pixelwise} and ACMP \cite{xu2020planar}.
We use the Adam optimizer \cite{kingma2014adam} with a learning rate of 1e-3 for network training, and set $\lambda_\text{\tiny{MVSRegu}}, \lambda_\text{\tiny{Smooth}}, \lambda_\text{\tiny{Eik}}$ to 0.5, 0.01, 0.03, respectively. 
For each iteration, we sample 5000 rays to train the model and use customized CUDA kernels for calculating the $\alpha$-compositing colors of the sampled points along each ray as \cref{eq:render_color}.
To maintain dynamic occupancy grids, we update the grids after every 16 training iterations and cap the mean density of grids by 1e-2 as the density threshold $\tau_\text{\tiny{occu}}$.

\noindent{\textbf{Evaluation Metrics}}
For 3D reconstruction, we assess the reconstructed surfaces in terms of Accuracy, Completeness, Precision, Recall, and F-score.
To evaluate the MVS predictions, we compute the distance differences for depth maps and count the angle errors for normal maps.
Please refer to the supplementary for more details about these evaluation metrics.

\begin{table}
    \centering
    \scalebox{0.66}{
    \begin{tabular}{c|lllll|c}
    \hline
    Method & Acc$\downarrow$ & Comp$\downarrow$ & Prec$\uparrow$ & Recall$\uparrow$ & F-score$\uparrow$ & Time$\downarrow$ \\
    \hline
    & & & & & & \\[-1em]
    COLMAP\cite{schonberger2016pixelwise} & 0.047 \markbronze & 0.235 & 0.711 & 0.441 & 0.537 
    & \textit{133} \\
    & & & & & & \\[-1em]
    ACMP\cite{xu2020planar} & 0.118 & 0.081 & 0.531 & 0.581 & 0.555 
    & \textit{10} \\
    & & & & & & \\[-1em]
    \hline
    & & & & & & \\[-1em]
    NeRF\cite{mildenhall2020nerf} & 0.735 & 0.177 & 0.131 & 0.290 & 0.176 
    & $>1$k \\
    & & & & & & \\[-1em]
    VolSDF\cite{yariv2021volume} & 0.414 & 0.120 & 0.321 & 0.394 & 0.346 
    & 825 \\
    & & & & & & \\[-1em]
    NeuS\cite{wang2021neus} & 0.179 & 0.208 & 0.313 & 0.275 & 0.291 
    & 531 \\
    & & & & & & \\[-1em]
    \hline
    & & & & & & \\[-1em]
    $\text{Manhattan-SDF}^\dag$\cite{guo2022neural} & 0.053 & 0.056 &
    0.715 \markbronze & 0.664 & 0.688
    & 528 \\
    & & & & & & \\[-1em]

    $\text{NeuRIS}^\dag$\cite{wang2022neuris} & 0.050 &
    0.049 \markbronze & 0.714 &
    0.670 \markbronze &
    0.691 \markbronze 
    & 406 \\

    & & & & & & \\[-1em]
    $\text{MonoSDF}^\dag$\cite{Yu2022MonoSDF} &
    0.035 \markgold &
    0.048 \marksilver &
    0.799 \markgold &
    0.681 \marksilver &
    0.733 \marksilver 
    & 708 \\

    & & & & & & \\[-1em]
    \hline
    & & & & & & \\[-1em]
    \textbf{HelixSurf}  &
    0.038 \marksilver &
    0.044 \markgold &
    0.786 \marksilver &
    0.727 \markgold &
    0.755 \markgold 
    & \textbf{33} \\
    \hline
    \end{tabular}
    }
    \caption{\textbf{Reconstruction metrics comparisons on ScanNet\cite{dai2017scannet}}. We compare our method with the state-of-the-art neural implicit surface learning methods\cite{mildenhall2020nerf, wang2021neus, yariv2021volume, guo2022neural, wang2022neuris, Yu2022MonoSDF} and PatchMatch based multi-view stereo methods (PM-MVS)\cite{schonberger2016pixelwise, xu2020planar}.
    Methods marked with $\dag$ are assisted with auxiliary training data, and vice versa.
    We mark the methods performing with least error using gold \protect\markgold, silver \protect\marksilver, and bronze \protect\markbronze \ medals.
    The last column shows time consumption (in minutes) for {\textit{PM-MVS}} methods, {Nueral implicit surface learning} methods, and {\textbf{HelixSurf}}.
    Note that the time for HelixSurf includes both MVS inference and neural implicit surface learning.
    }
    \label{table:benchmark}
\end{table}

\begin{figure*}
    \centering
    \begin{minipage}[t]{1.0\linewidth}
        \centering
        \includegraphics[width=\textwidth]{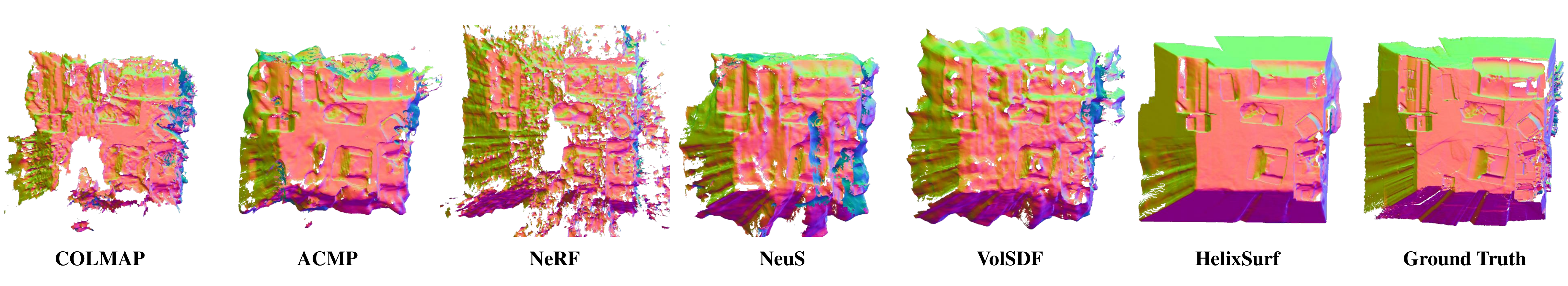}
        \subcaption{Comparisons with existing methods when these methods do not use auxiliary training data. }
        \label{fig:comparison1}
    \end{minipage} \quad
    \begin{minipage}[t]{1.0\linewidth}
        \includegraphics[width=\textwidth]{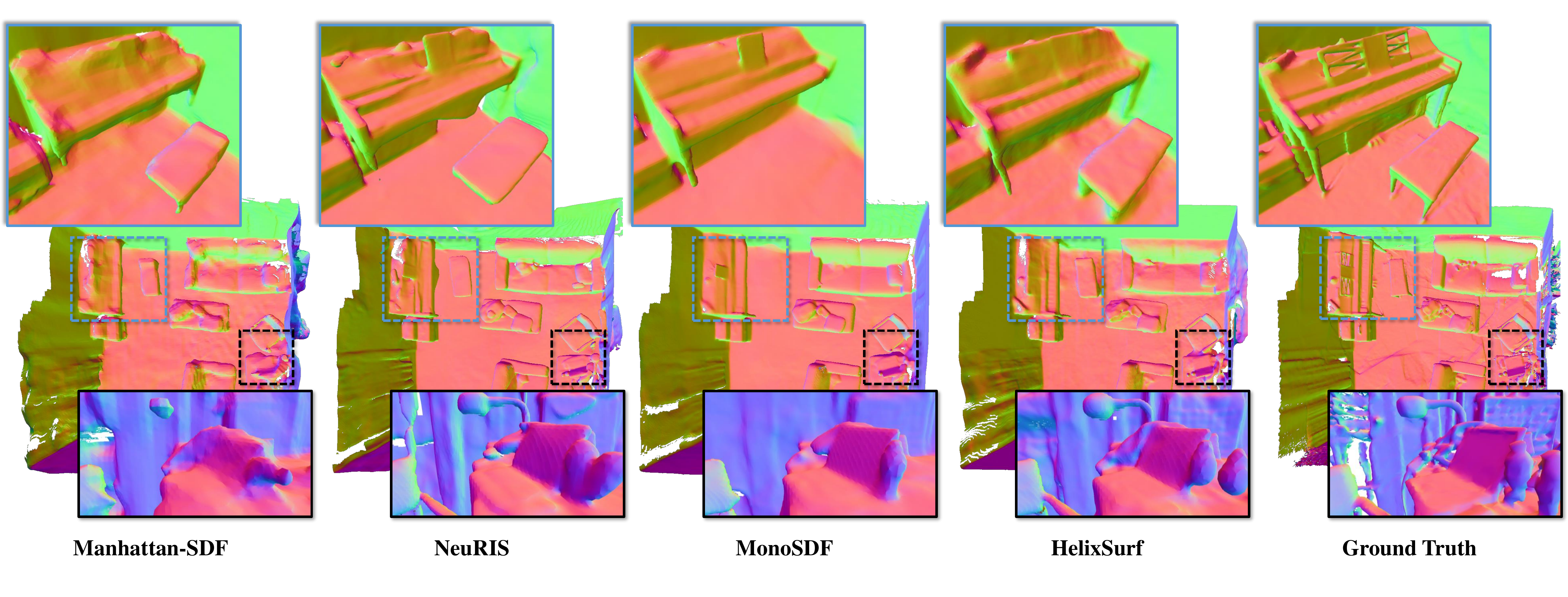}
        \subcaption{Comparisons with existing methods when they invoke the use of auxiliary training data. Note that our HelixSurf do not use auxiliary training data. }
        \label{fig:compariso2}
    \end{minipage}
    \caption{
    \textbf{Qualitative geometry comparisons on ScanNet.} Compared to existing methods, our method can better reconstruct the scene details (\eg the lamp, the cabinet and chair handles) and the smooth regions (\eg the floor and walls). Surface normals are visualized as coded colors.
    }
    \label{fig:comparison}
\end{figure*}

\subsection{Comparisons}
\label{subsec:comparisons}
We evaluate the 3D geometry metrics and time consumption of our proposed HelixSurf against existing methods on ScanNet \cite{dai2017scannet}, as shown in \cref{table:benchmark}. 
Each quantitative result is averaged over all the selected scenes.
For the geometry comparison, HelixSurf manifestly surpasses existing methods in almost every metrics, even some of the comparison methods are assisted with auxiliary training data. And the qualitative results in \cref{fig:comparison} further support the quantitative analyses.
Without auxiliary training data, HelixSurf is capable of handling the textureless surface areas where other methods fail to tackle, as in \cref{fig:comparison}(a).
Moreover, HelixSurf produces better details of objects than those methods using auxiliary training data, as in \cref{fig:comparison}(b).
As for learning time, the data in \cref{table:benchmark} indicates that HelixSurf improves the learning efficiency with orders of magnitude when compared with existing neural implicit surface learning methods, even with the inclusion of MVS inference time.

\subsection{Ablation Studies}
\label{subsec:ablation}
HelixSurf is optimized with interactive intertwined regularization as stated in Section \ref{sec:methods}.
We design elaborate experiments to evaluate the efficacy of this regularization.
Furthermore, the sampling guided by dynamic occupancy grids (\cf Section \ref{subsec:occupancy_grids}) is essential to realize fast training convergence. 
We thus compare it with the ordinary sampling alternative. 
These studies are conducted on the ScanNet dataset \cite{dai2017scannet}.

\begin{figure}[ht]
    \centering
    \begin{minipage}[t]{0.43\linewidth}
        \centering
        \includegraphics[width=\textwidth]{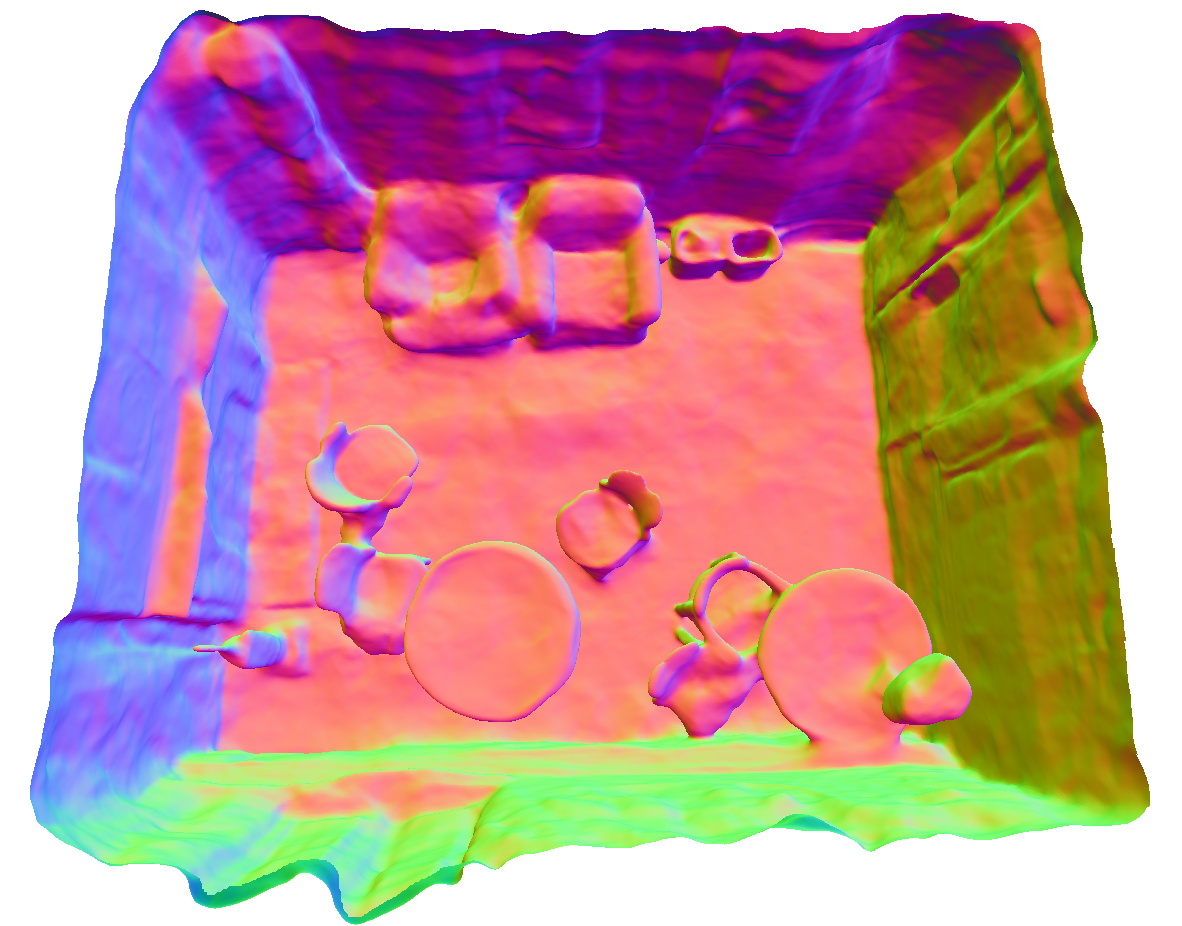}
        \subcaption{W/O smoothness on textureless surface areas}
        \label{fig:non_planar}
    \end{minipage} \quad
    \begin{minipage}[t]{0.43\linewidth}
        \includegraphics[width=\textwidth]{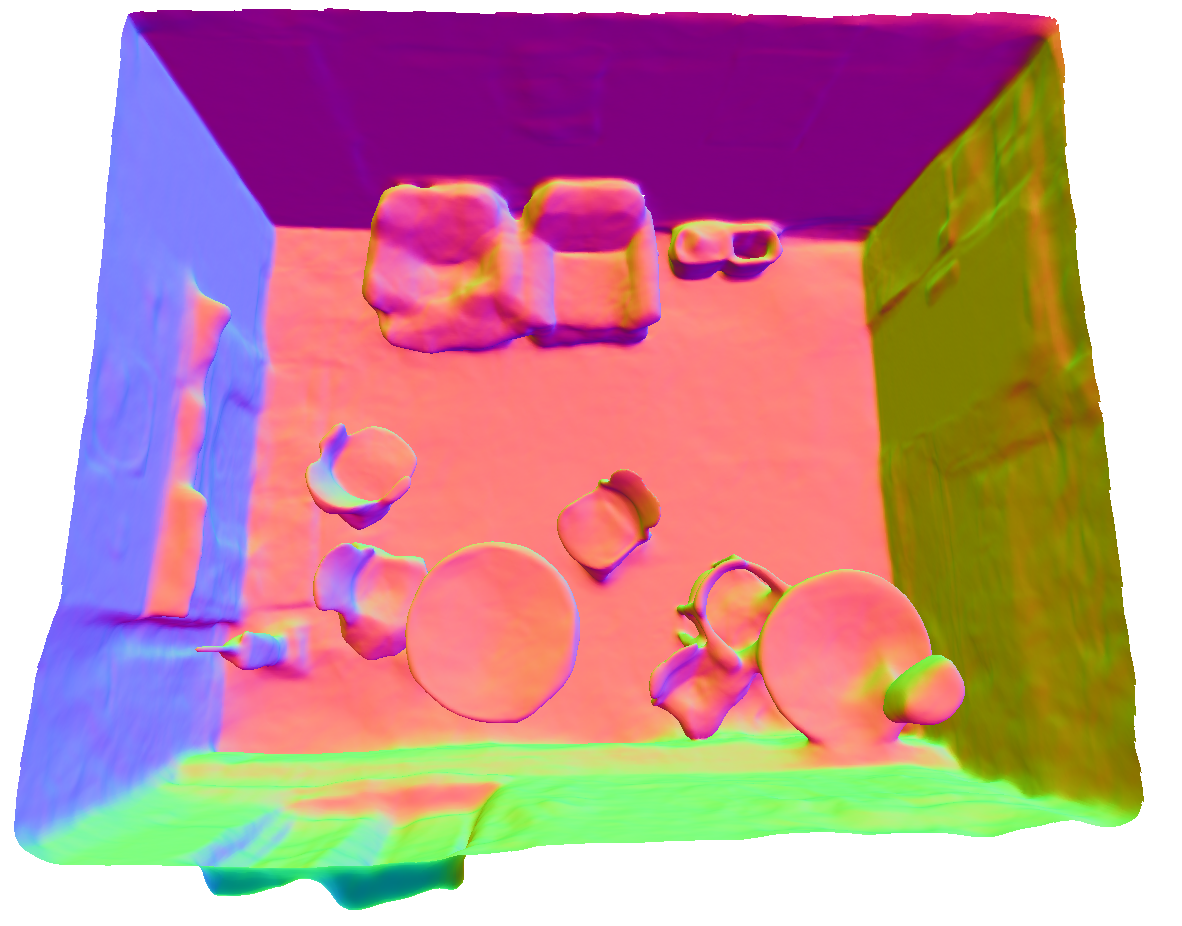}
        \subcaption{With smoothness on textureless surface areas}
        \label{fig:add_planar}
    \end{minipage}
    \caption{
    Visualization of example reconstruction without the use of smoothness scheme for textureless areas (a) and with the use of smoothness scheme (b). The colors encode surface normals.
    }
    \label{fig:ablation_planar}
\end{figure}

\noindent{\textbf{Analysis on the regularization of neural implicit surface learning from MVS predictions }}
The MVS inference results effectively regularize the neural implicit surface learning (\cf Section \ref{subsec:reg_neural_from_MVS}) and facilitate the network to capture fine details.
The results in \cref{table:ablation} illustrate that the MVS predictions effectively promote the surface learning and the \emph{regularized} MVS predictions can further improve the quality of reconstruction.
Nonetheless, the MVS predictions are less reliable on the textureless surface areas, we thus leverage the homogeneity inside individual superpixels and devise a scheme (\cf Section \ref{subsec:handling_textureless_areas}) to regularize the learning on such areas.
As shown in \cref{fig:ablation_planar}, our proposed scheme handles the textureless surface areas and reconstructs smoother surface on the basis of maintaining the details of non-planar regions.

\begin{figure}[ht]
    \centering
    \begin{minipage}[t]{0.49\linewidth}
        \centering
        \includegraphics[width=\textwidth]{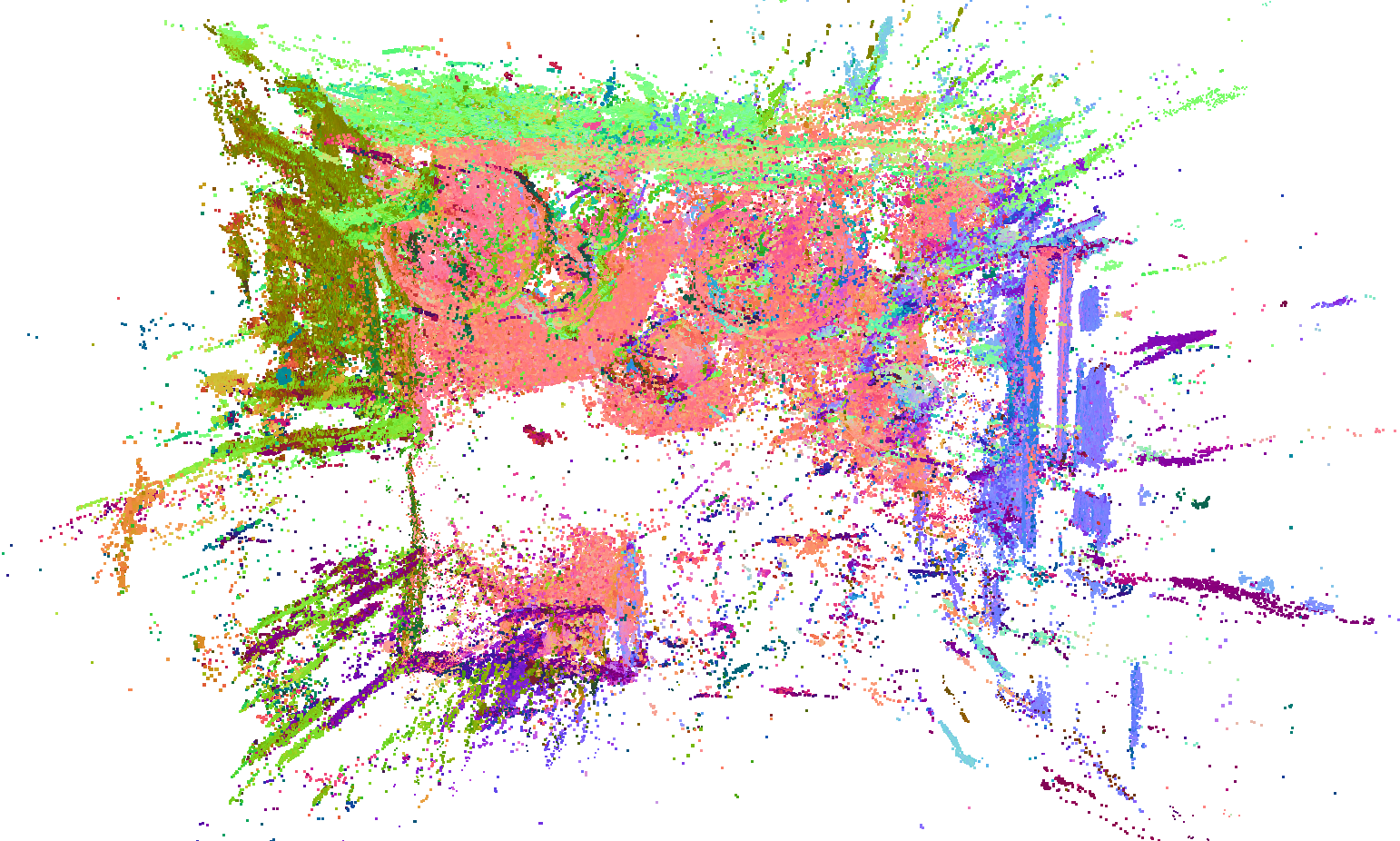}
        \subcaption{ordinary MVS.}
        \label{fig:uniform}
    \end{minipage}
    \begin{minipage}[t]{0.49\linewidth}
        \includegraphics[width=\textwidth]{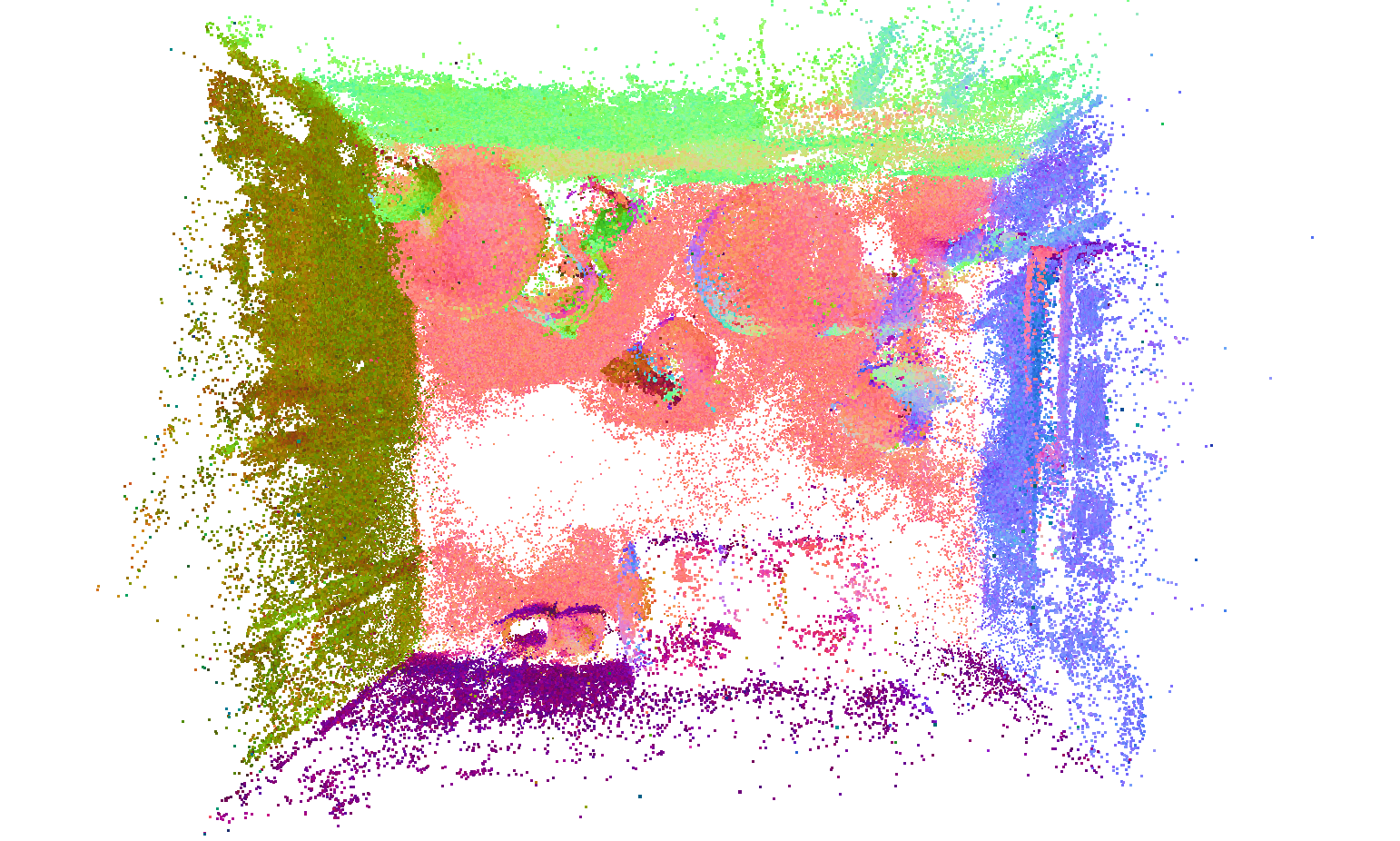}
        \subcaption{regularized MVS.}
        \label{fig:improved}
    \end{minipage}
    \caption{Qualitative comparisons of inference results between the ordinary MVS method (a) and our regularized MVS (b).}
    \label{fig:improved_geometric}
\end{figure}

\noindent{\textbf{Analysis on the regularization of MVS from neural implicit surface learning}}
During the training process of the neural surface, the underlying geometries are progressively recovered.
We use the learned depths and normals as priors to regularize MVS, which clears up the artifacts produced by the ordinary MVS and enables the double helix to forward and rise.
\cref{fig:improved_geometric} qualitatively shows the inference results of the ordinary MVS method (\cref{fig:improved_geometric}(a)) and the regularized MVS (\cref{fig:improved_geometric}(b)). 
\cref{table:mvs_comparison} shows the quantitative comparison between the ordinary MVS method and our regularized one. 
Both qualitative and quantitative comparisons verify that this regularization eliminates noise and outliers and improves the quality of inference results.

\noindent{\textbf{Efficacy of Sampling Guided by Dynamic Occupancy Grids}}
To further alleviate the difficulties of optimization, we maintain dynamic occupancy grids and propose a sampling strategy to skip the sample points in empty space.
The time consumption comparisons in \cref{table:benchmark} show that our training convergence is significantly faster than the existing neural implicit surface learning methods.
The results in \cref{table:traintime_composition} present the timing consumptions of each part in the entire training process with and without the sampling strategy, respectively.

\begin{table}[!htb]
    \centering
    \scalebox{0.61}{
    \begin{tabular}{ccc|ccccc}
    \hline
    \multicolumn{3}{c|}{Regularization} &
    \multirow{3}{*}{Acc$\downarrow$} &
    \multirow{3}{*}{Comp$\downarrow$} &
    \multirow{3}{*}{Prec$\uparrow$} &
    \multirow{3}{*}{Recall$\uparrow$} &
    \multirow{3}{*}{F-score$\uparrow$} \\
    \cline{1-3}
    \makecell{oridinary\\MVS} &
    \makecell{regularized\\MVS}&
    \makecell{Textureless\\Areas Handling} & \\
    \hline

    & & & & & & & \\[-0.8em]
    & & & 0.179 & 0.208 & 0.313 & 0.275 & 0.291 \\

    & & & & & & & \\[-0.8em]
    \checkmark & & & 
    0.059 & 0.076 & 0.661 & 0.605 & 0.632 \\
    
    & & & & & & & \\[-0.8em]
    & \checkmark & & 
    0.051 & 0.066 & 0.711 & 0.649 & 0.679 \\
    
    & & & & & & & \\[-0.8em]
    \checkmark & & \checkmark &
    0.047 & 0.053 & 0.768 & 0.706 & 0.735 \\
    
    & & & & & & & \\[-0.8em]
    & \checkmark & \checkmark &
    \textbf{0.038} & \textbf{0.044} & \textbf{0.786} & \textbf{0.727} & \textbf{0.755} \\
    
    \hline
    \end{tabular}
    }
    \caption{Analyses on the regularization of neural implicit surface learning from MVS predictions.}
    \label{table:ablation}
\end{table}

\begin{table}[!htb]
    \centering
    \scalebox{0.9}{
    \begin{tabular}{c|cccc}
    \hline
    \multirow{2}{*}{\makecell{Method}} & \multicolumn{4}{c}{Depth map} \\
    \cline{2-5}
    & Abs Diff$\downarrow$ & Abs Rel$\downarrow$ & Sq Rel$\downarrow$ & RMSE $\downarrow$\\
    \hline
    & & & & \\[-0.9em]
    $\text{ordinary}$ & 0.067 & 0.098 & 0.020 & 0.147 \\
    & & & & \\[-0.9em]
    $\text{regularized}$     & 0.053 & 0.085 & 0.011 & 0.106 \\
    \hline
    
    \multirow{2}{*}{\makecell{Method}} & \multicolumn{4}{c}{Normal map} \\
    \cline{2-5}
    & Mean $\downarrow$ & Median$\downarrow$ & RMSE$\downarrow$ & $\text{Prop}\_30^{\circ}\uparrow$ \\
    \hline
    & & & & \\[-0.9em]
    $\text{ordinary}$ & 35.5$^{\circ}$ & 30.4$^{\circ}$ & 42.6$^{\circ}$ & 51.0\%  \\
    & & & & \\[-0.9em]
    $\text{regularized}$     & 27.8$^{\circ}$ & 20.2$^{\circ}$ & 35.3$^{\circ}$ & 67.4\% \\
    \hline
    \end{tabular}
    }
    \caption{
    Quantitative comparison between the ordinary MVS and our regularized MVS.
    }
    \label{table:mvs_comparison}
\end{table}

\begin{table}[!htb]
    \centering
    \scalebox{0.85}{
    \begin{tabular}{l|c|c|c|c|c|c}
    \hline
   \makecell{Occ\\Grids} & MVS & \makecell{Texture-\\less} & Grid & \makecell{Training\\Forward} & \makecell{Training\\Backward} & \textbf{Total} \\
    \hline
    w/ & \multirow{2}{*}{3.8} & \multirow{2}{*}{2.6} & 0.6 & 12.4 & 13.8 & \textbf{33.2} \\
    \cline{1-1}\cline{4-7}
    % \hline
    w/o &  &  & - & 184 & 203 & 393.4 \\
    \hline
    \end{tabular}
    }
    \caption{
    Time consumption (in minutes) for each part of our training process with or without guidance of Dynamic Occupancy Grids.
    }
    \label{table:traintime_composition}
\end{table}

\begin{figure}[htbp]
    \centering
    \begin{minipage}[t]{0.32\linewidth}
        \centering
        \includegraphics[width=\textwidth]{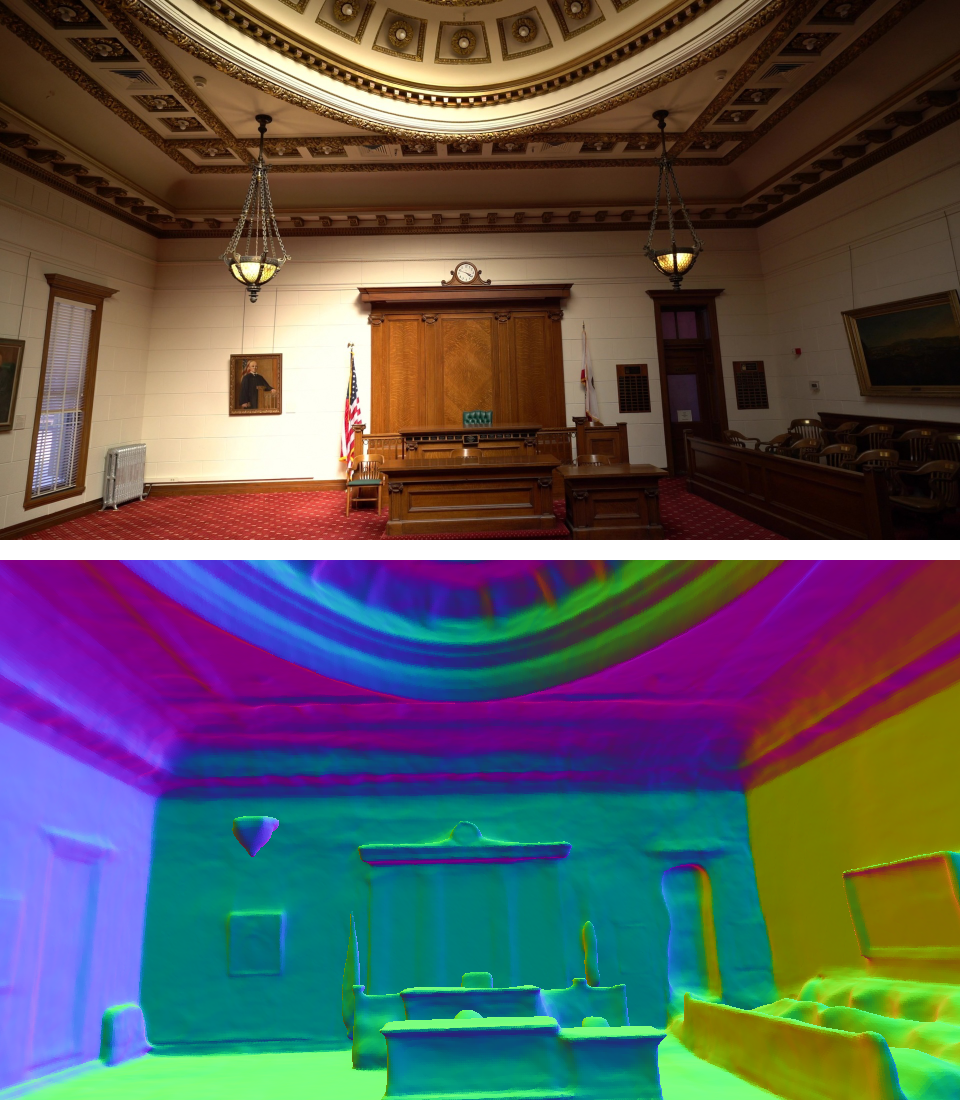}
        \subcaption{}
        \label{fig:indoor1}
    \end{minipage}
    \begin{minipage}[t]{0.32\linewidth}
        \includegraphics[width=\textwidth]{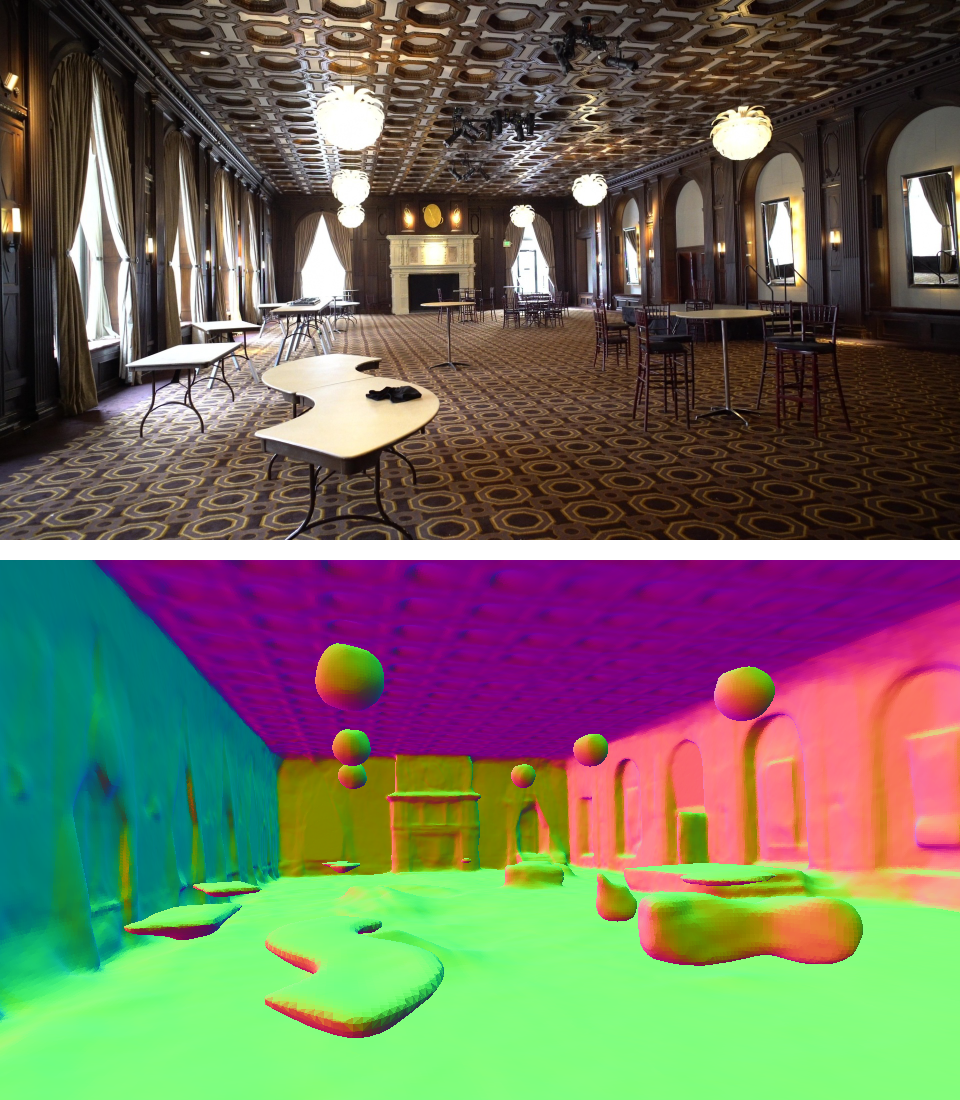}
        \subcaption{}
        \label{fig:indoor2}
    \end{minipage}
    \begin{minipage}[t]{0.32\linewidth}
        \includegraphics[width=\textwidth]{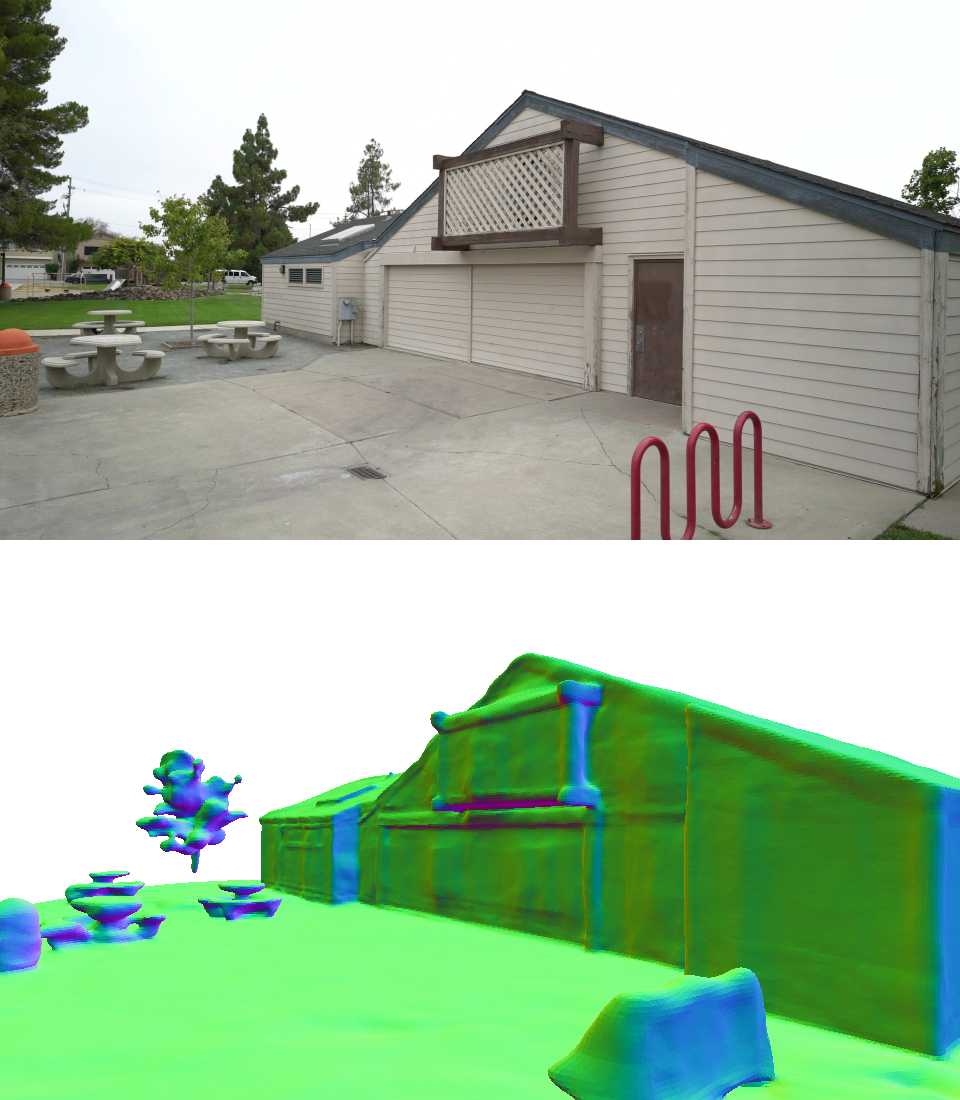}
        \subcaption{}
        \label{fig:outdoor}
    \end{minipage}
    \caption{\textbf{Qualitative result of the reconstruction on Tanks and Temples \cite{knapitsch2017tanks}.} (a) and (b) are examples of indoor scenes. (c) is an outdoor scene.}
    \label{fig:TnT}
\end{figure}

\subsection{Real-world Large-scale Scene Reconstruction}
To further examine the applicability and generalization of HelixSurf, we conduct experiments on an indoor subset from the Tanks and Temples \cite{knapitsch2017tanks} dataset.
Results in \cref{fig:TnT}(a, b) show that HelixSurf achieves reasonable results on such large-scale indoor scenes.
Furthermore, we evaluate HelixSurf on large-scale outdoor scenes from Tanks and Temples \cite{knapitsch2017tanks}.
Surprisingly, HelixSurf has potential to handle large-scale outdoor scenes, as shown in \cref{fig:TnT}(c).
Please refer to the supplementary for more results.

%%%%%%%%% REFERENCES
{\small
\bibliographystyle{ieee_fullname}
\bibliography{egbib}

\begin{thebibliography}{10}\itemsep=-1pt

\bibitem{aanaes2016large}
Henrik Aan{\ae}s, Rasmus~Ramsb{\o}l Jensen, George Vogiatzis, Engin Tola, and
  Anders~Bjorholm Dahl.
\newblock Large-scale data for multiple-view stereopsis.
\newblock {\em International Journal of Computer Vision}, 120(2):153--168,
  2016.

\bibitem{achanta2012slic}
Radhakrishna Achanta, Appu Shaji, Kevin Smith, Aurelien Lucchi, Pascal Fua, and
  Sabine S{\"u}sstrunk.
\newblock Slic superpixels compared to state-of-the-art superpixel methods.
\newblock {\em IEEE transactions on pattern analysis and machine intelligence},
  34(11):2274--2282, 2012.

\bibitem{barnes2009patchmatch}
Connelly Barnes, Eli Shechtman, Adam Finkelstein, and Dan~B Goldman.
\newblock Patchmatch: A randomized correspondence algorithm for structural
  image editing.
\newblock {\em ACM Trans. Graph.}, 28(3):24, 2009.

\bibitem{bleyer2011patchmatch}
Michael Bleyer, Christoph Rhemann, and Carsten Rother.
\newblock Patchmatch stereo-stereo matching with slanted support windows.
\newblock In {\em Bmvc}, volume~11, pages 1--11, 2011.

\bibitem{chen20153survey}
Kang Chen, Yu-Kun Lai, and Shi-Min Hu.
\newblock 3d indoor scene modeling from rgb-d data: a survey.
\newblock {\em Computational Visual Media}, 1(4):267--278, 2015.

\bibitem{chibane2020neural}
Julian Chibane, Gerard Pons-Moll, et~al.
\newblock Neural unsigned distance fields for implicit function learning.
\newblock {\em Advances in Neural Information Processing Systems},
  33:21638--21652, 2020.

\bibitem{dai2017scannet}
Angela Dai, Angel~X Chang, Manolis Savva, Maciej Halber, Thomas Funkhouser, and
  Matthias Nie{\ss}ner.
\newblock Scannet: Richly-annotated 3d reconstructions of indoor scenes.
\newblock In {\em Proceedings of the IEEE conference on computer vision and
  pattern recognition}, pages 5828--5839, 2017.

\bibitem{dai2017bundlefusion}
Angela Dai, Matthias Nie{\ss}ner, Michael Zollh{\"o}fer, Shahram Izadi, and
  Christian Theobalt.
\newblock Bundlefusion: Real-time globally consistent 3d reconstruction using
  on-the-fly surface reintegration.
\newblock {\em ACM Transactions on Graphics (ToG)}, 36(4):1, 2017.

\bibitem{darmon2022improving}
Fran{\c{c}}ois Darmon, B{\'e}n{\'e}dicte Bascle, Jean-Cl{\'e}ment Devaux,
  Pascal Monasse, and Mathieu Aubry.
\newblock Improving neural implicit surfaces geometry with patch warping.
\newblock In {\em Proceedings of the IEEE/CVF Conference on Computer Vision and
  Pattern Recognition}, pages 6260--6269, 2022.

\bibitem{felzenszwalb2004efficient}
Pedro~F Felzenszwalb and Daniel~P Huttenlocher.
\newblock Efficient graph-based image segmentation.
\newblock {\em International journal of computer vision}, 59(2):167--181, 2004.

\bibitem{fu2022geo}
Qiancheng Fu, Qingshan Xu, Yew-Soon Ong, and Wenbing Tao.
\newblock Geo-neus: Geometry-consistent neural implicit surfaces learning for
  multi-view reconstruction.
\newblock {\em arXiv preprint arXiv:2205.15848}, 2022.

\bibitem{furukawa2015multi}
Yasutaka Furukawa, Carlos Hern{\'a}ndez, et~al.
\newblock Multi-view stereo: A tutorial.
\newblock {\em Foundations and Trends{\textregistered} in Computer Graphics and
  Vision}, 9(1-2):1--148, 2015.

\bibitem{galliani2015massively}
Silvano Galliani, Katrin Lasinger, and Konrad Schindler.
\newblock Massively parallel multiview stereopsis by surface normal diffusion.
\newblock In {\em Proceedings of the IEEE International Conference on Computer
  Vision}, pages 873--881, 2015.

\bibitem{gropp2020implicit}
Amos Gropp, Lior Yariv, Niv Haim, Matan Atzmon, and Yaron Lipman.
\newblock Implicit geometric regularization for learning shapes.
\newblock In {\em Proceedings of the 37th International Conference on Machine
  Learning}, pages 3789--3799, 2020.

\bibitem{guo2022neural}
Haoyu Guo, Sida Peng, Haotong Lin, Qianqian Wang, Guofeng Zhang, Hujun Bao, and
  Xiaowei Zhou.
\newblock Neural 3d scene reconstruction with the manhattan-world assumption.
\newblock In {\em Proceedings of the IEEE/CVF Conference on Computer Vision and
  Pattern Recognition}, pages 5511--5520, 2022.

\bibitem{mvg_book}
Richard Hartley and Andrew Zisserman.
\newblock {\em Multiple view geometry in computer vision}.
\newblock Cambridge university press, 2003.

\bibitem{huang2022surface}
Zhangjin Huang, Yuxin Wen, Zihao Wang, Jinjuan Ren, and Kui Jia.
\newblock Surface reconstruction from point clouds: A survey and a benchmark.
\newblock {\em arXiv preprint arXiv:2205.02413}, 2022.

\bibitem{im2018dpsnet}
Sunghoon Im, Hae-Gon Jeon, Stephen Lin, and In~So Kweon.
\newblock Dpsnet: End-to-end deep plane sweep stereo.
\newblock In {\em International Conference on Learning Representations}, 2018.

\bibitem{kazhdan2006poisson}
Michael Kazhdan, Matthew Bolitho, and Hugues Hoppe.
\newblock Poisson surface reconstruction.
\newblock In {\em Proceedings of the fourth Eurographics symposium on Geometry
  processing}, volume~7, 2006.

\bibitem{kazhdan2013screened}
Michael Kazhdan and Hugues Hoppe.
\newblock Screened poisson surface reconstruction.
\newblock {\em ACM Transactions on Graphics (ToG)}, 32(3):1--13, 2013.

\bibitem{kingma2014adam}
Diederik~P Kingma and Jimmy Ba.
\newblock Adam: A method for stochastic optimization.
\newblock {\em arXiv preprint arXiv:1412.6980}, 2014.

\bibitem{knapitsch2017tanks}
Arno Knapitsch, Jaesik Park, Qian-Yi Zhou, and Vladlen Koltun.
\newblock Tanks and temples: Benchmarking large-scale scene reconstruction.
\newblock {\em ACM Transactions on Graphics (ToG)}, 36(4):1--13, 2017.

\bibitem{labatut2007efficient}
Patrick Labatut, Jean-Philippe Pons, and Renaud Keriven.
\newblock Efficient multi-view reconstruction of large-scale scenes using
  interest points, delaunay triangulation and graph cuts.
\newblock In {\em 2007 IEEE 11th international conference on computer vision},
  pages 1--8. IEEE, 2007.

\bibitem{lei2020analytic}
Jiabao Lei and Kui Jia.
\newblock Analytic marching: An analytic meshing solution from deep implicit
  surface networks.
\newblock In {\em International Conference on Machine Learning}, pages
  5789--5798. PMLR, 2020.

\bibitem{liu2020dist}
Shaohui Liu, Yinda Zhang, Songyou Peng, Boxin Shi, Marc Pollefeys, and Zhaopeng
  Cui.
\newblock Dist: Rendering deep implicit signed distance function with
  differentiable sphere tracing.
\newblock In {\em Proceedings of the IEEE/CVF Conference on Computer Vision and
  Pattern Recognition}, pages 2019--2028, 2020.

\bibitem{lorensen1987marching}
William~E Lorensen and Harvey~E Cline.
\newblock Marching cubes: A high resolution 3d surface construction algorithm.
\newblock {\em ACM siggraph computer graphics}, 21(4):163--169, 1987.

\bibitem{max1995optical}
Nelson Max.
\newblock Optical models for direct volume rendering.
\newblock {\em IEEE Transactions on Visualization and Computer Graphics},
  1(2):99--108, 1995.

\bibitem{mescheder2019occupancy}
Lars Mescheder, Michael Oechsle, Michael Niemeyer, Sebastian Nowozin, and
  Andreas Geiger.
\newblock Occupancy networks: Learning 3d reconstruction in function space.
\newblock In {\em Proceedings of the IEEE/CVF conference on computer vision and
  pattern recognition}, pages 4460--4470, 2019.

\bibitem{mildenhall2020nerf}
Ben Mildenhall, Pratul~P Srinivasan, Matthew Tancik, Jonathan~T Barron, Ravi
  Ramamoorthi, and Ren Ng.
\newblock Nerf: Representing scenes as neural radiance fields for view
  synthesis.
\newblock In {\em Computer Vision--ECCV 2020: 16th European Conference,
  Glasgow, UK, August 23--28, 2020, Proceedings, Part I}, pages 405--421, 2020.

\bibitem{niemeyer2020differentiable}
Michael Niemeyer, Lars Mescheder, Michael Oechsle, and Andreas Geiger.
\newblock Differentiable volumetric rendering: Learning implicit 3d
  representations without 3d supervision.
\newblock In {\em Proceedings of the IEEE/CVF Conference on Computer Vision and
  Pattern Recognition}, pages 3504--3515, 2020.

\bibitem{oechsle2021unisurf}
Michael Oechsle, Songyou Peng, and Andreas Geiger.
\newblock Unisurf: Unifying neural implicit surfaces and radiance fields for
  multi-view reconstruction.
\newblock In {\em Proceedings of the IEEE/CVF International Conference on
  Computer Vision}, pages 5589--5599, 2021.

\bibitem{park2019deepsdf}
Jeong~Joon Park, Peter Florence, Julian Straub, Richard Newcombe, and Steven
  Lovegrove.
\newblock Deepsdf: Learning continuous signed distance functions for shape
  representation.
\newblock In {\em Proceedings of the IEEE/CVF conference on computer vision and
  pattern recognition}, pages 165--174, 2019.

\bibitem{parker2010optix}
Steven~G Parker, James Bigler, Andreas Dietrich, Heiko Friedrich, Jared
  Hoberock, David Luebke, David McAllister, Morgan McGuire, Keith Morley,
  Austin Robison, et~al.
\newblock Optix: a general purpose ray tracing engine.
\newblock {\em Acm transactions on graphics (tog)}, 29(4):1--13, 2010.

\bibitem{paszke2019pytorch}
Adam Paszke, Sam Gross, Francisco Massa, Adam Lerer, James Bradbury, Gregory
  Chanan, Trevor Killeen, Zeming Lin, Natalia Gimelshein, Luca Antiga, et~al.
\newblock Pytorch: An imperative style, high-performance deep learning library.
\newblock {\em Advances in neural information processing systems}, 32, 2019.

\bibitem{peng2020convolutional}
Songyou Peng, Michael Niemeyer, Lars Mescheder, Marc Pollefeys, and Andreas
  Geiger.
\newblock Convolutional occupancy networks.
\newblock In {\em European Conference on Computer Vision}, pages 523--540.
  Springer, 2020.

\bibitem{romanoni2019tapa}
Andrea Romanoni and Matteo Matteucci.
\newblock Tapa-mvs: Textureless-aware patchmatch multi-view stereo.
\newblock In {\em Proceedings of the IEEE/CVF International Conference on
  Computer Vision}, pages 10413--10422, 2019.

\bibitem{schonberger2016structure}
Johannes~L Schonberger and Jan-Michael Frahm.
\newblock Structure-from-motion revisited.
\newblock In {\em Proceedings of the IEEE conference on computer vision and
  pattern recognition}, pages 4104--4113, 2016.

\bibitem{schonberger2016pixelwise}
Johannes~L Sch{\"o}nberger, Enliang Zheng, Jan-Michael Frahm, and Marc
  Pollefeys.
\newblock Pixelwise view selection for unstructured multi-view stereo.
\newblock In {\em European conference on computer vision}, pages 501--518.
  Springer, 2016.

\bibitem{shen2013accurate}
Shuhan Shen.
\newblock Accurate multiple view 3d reconstruction using patch-based stereo for
  large-scale scenes.
\newblock {\em IEEE transactions on image processing}, 22(5):1901--1914, 2013.

\bibitem{wang2021patchmatchnet}
Fangjinhua Wang, Silvano Galliani, Christoph Vogel, Pablo Speciale, and Marc
  Pollefeys.
\newblock Patchmatchnet: Learned multi-view patchmatch stereo.
\newblock In {\em Proceedings of the IEEE/CVF Conference on Computer Vision and
  Pattern Recognition}, pages 14194--14203, 2021.

\bibitem{wang2022neuris}
Jiepeng Wang, Peng Wang, Xiaoxiao Long, Christian Theobalt, Taku Komura,
  Lingjie Liu, and Wenping Wang.
\newblock Neuris: Neural reconstruction of indoor scenes using normal priors.
\newblock {\em arXiv preprint arXiv:2206.13597}, 2022.

\bibitem{wang2021neus}
Peng Wang, Lingjie Liu, Yuan Liu, Christian Theobalt, Taku Komura, and Wenping
  Wang.
\newblock Neus: Learning neural implicit surfaces by volume rendering for
  multi-view reconstruction.
\newblock {\em Advances in Neural Information Processing Systems},
  34:27171--27183, 2021.

\bibitem{wang2022hfneus}
Yiqun Wang, Ivan Skorokhodov, and Peter Wonka.
\newblock Hf-neus: Improved surface reconstruction using high-frequency
  details.
\newblock {\em arXiv preprint arXiv:2206.07850}, 2022.

\bibitem{xu2018multi}
Qingshan Xu and Wenbing Tao.
\newblock Multi-view stereo with asymmetric checkerboard propagation and
  multi-hypothesis joint view selection.
\newblock {\em arXiv preprint arXiv:1805.07920}, 2018.

\bibitem{Xu2019ACMM}
Qingshan Xu and Wenbing Tao.
\newblock Multi-scale geometric consistency guided multi-view stereo.
\newblock In {\em Proceedings of the IEEE/CVF Conference on Computer Vision and
  Pattern Recognition (CVPR)}, June 2019.

\bibitem{xu2020planar}
Qingshan Xu and Wenbing Tao.
\newblock Planar prior assisted patchmatch multi-view stereo.
\newblock In {\em Proceedings of the AAAI Conference on Artificial
  Intelligence}, pages 12516--12523, 2020.

\bibitem{xu2020pvsnet}
Qingshan Xu and Wenbing Tao.
\newblock Pvsnet: Pixelwise visibility-aware multi-view stereo network.
\newblock {\em arXiv preprint arXiv:2007.07714}, 2020.

\bibitem{yao2018mvsnet}
Yao Yao, Zixin Luo, Shiwei Li, Tian Fang, and Long Quan.
\newblock Mvsnet: Depth inference for unstructured multi-view stereo.
\newblock In {\em Proceedings of the European conference on computer vision
  (ECCV)}, pages 767--783, 2018.

\bibitem{yao2019recurrent}
Yao Yao, Zixin Luo, Shiwei Li, Tianwei Shen, Tian Fang, and Long Quan.
\newblock Recurrent mvsnet for high-resolution multi-view stereo depth
  inference.
\newblock In {\em Proceedings of the IEEE/CVF conference on computer vision and
  pattern recognition}, pages 5525--5534, 2019.

\bibitem{yariv2021volume}
Lior Yariv, Jiatao Gu, Yoni Kasten, and Yaron Lipman.
\newblock Volume rendering of neural implicit surfaces.
\newblock {\em Advances in Neural Information Processing Systems},
  34:4805--4815, 2021.

\bibitem{yariv2020multiview}
Lior Yariv, Yoni Kasten, Dror Moran, Meirav Galun, Matan Atzmon, Basri Ronen,
  and Yaron Lipman.
\newblock Multiview neural surface reconstruction by disentangling geometry and
  appearance.
\newblock {\em Advances in Neural Information Processing Systems},
  33:2492--2502, 2020.

\bibitem{yu2021plenoctrees}
Alex Yu, Ruilong Li, Matthew Tancik, Hao Li, Ren Ng, and Angjoo Kanazawa.
\newblock Plenoctrees for real-time rendering of neural radiance fields.
\newblock In {\em Proceedings of the IEEE/CVF International Conference on
  Computer Vision}, pages 5752--5761, 2021.

\bibitem{Yu2022MonoSDF}
Zehao Yu, Songyou Peng, Michael Niemeyer, Torsten Sattler, and Andreas Geiger.
\newblock Monosdf: Exploring monocular geometric cues for neural implicit
  surface reconstruction.
\newblock {\em Advances in Neural Information Processing Systems (NeurIPS)},
  2022.

\bibitem{zhang2020nerf++}
Kai Zhang, Gernot Riegler, Noah Snavely, and Vladlen Koltun.
\newblock Nerf++: Analyzing and improving neural radiance fields.
\newblock {\em arXiv preprint arXiv:2010.07492}, 2020.

\bibitem{zheng2014patchmatch}
Enliang Zheng, Enrique Dunn, Vladimir Jojic, and Jan-Michael Frahm.
\newblock Patchmatch based joint view selection and depthmap estimation.
\newblock In {\em Proceedings of the IEEE Conference on Computer Vision and
  Pattern Recognition}, pages 1510--1517, 2014.

\end{thebibliography}
}

\newpage

\appendix

\vbox{
\centering
\Large{
\twocolumn[
    \centerline{\textbf{HelixSurf: A Robust and Efficient Neural Implicit Surface Learning of Indoor}}
    \centerline{\textbf{Scenes with Iterative Intertwined Regularization - Supplementary Material}}
    \vskip 0.2in
]}
}

%%%%%%%%% BODY TEXT
\section{Network Architecture}
\label{sec:architecture}
\begin{figure}[htbp]
    \centering
    \includegraphics[width=0.47\textwidth]{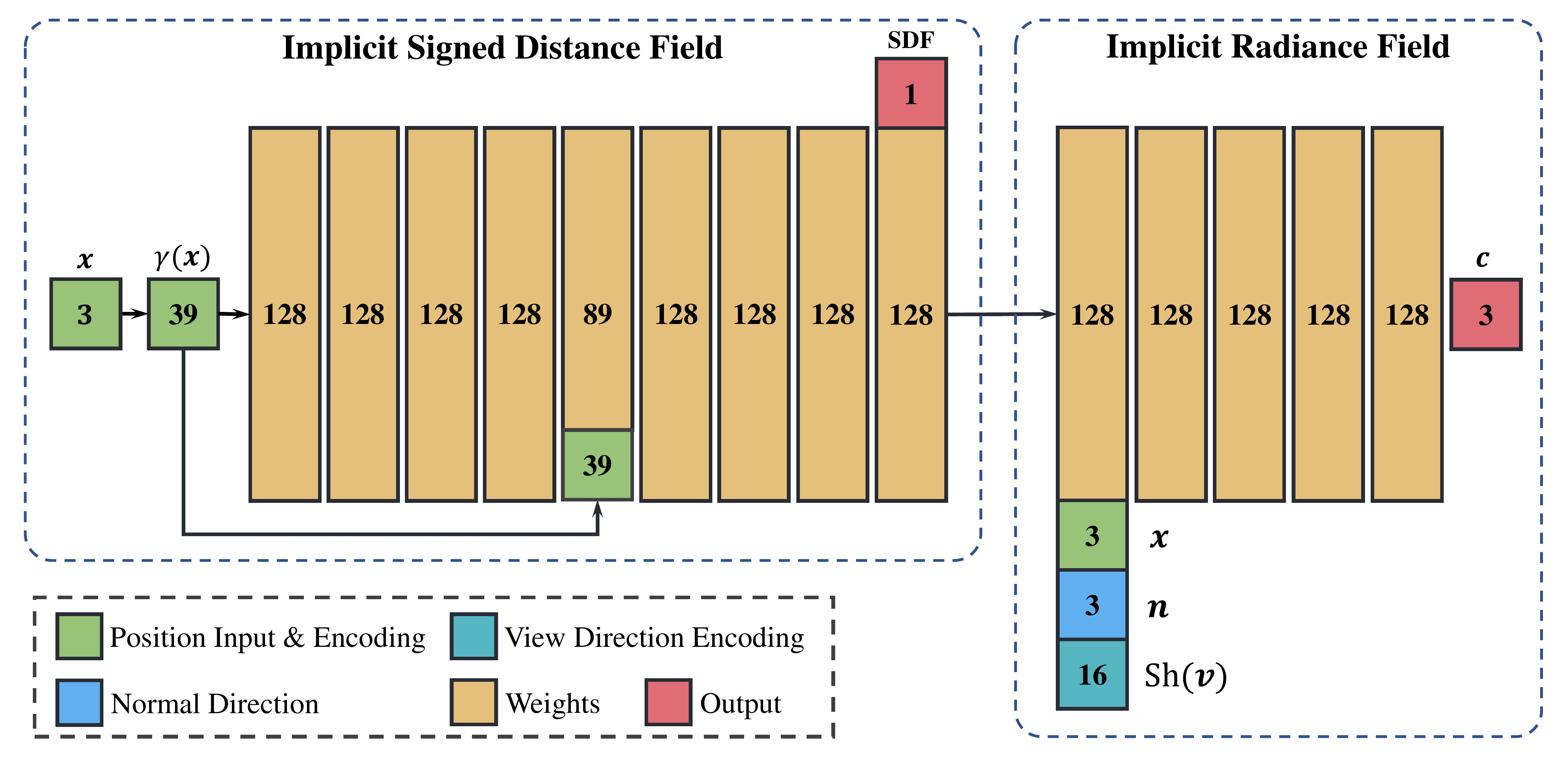}
    \vspace{-0.3cm}
    \caption{\textbf{Architecture of HelixSurf.} HelixSurf takes the position $\bm{x}$ and the view direction $\bm{v}$ of the sample point as inputs, and outputs the $\text{SDF}$ and color $\bm{c}$. The surface normal $\bm{n} = \nabla f(\bm{x})$.}
    \label{fig:architecture}
\end{figure}

HelixSurf uses two MLPs to encode the implicit signed distance field (SDF-MLP) and implicit radiance field (RF-MLP), respectively. 
The architecture of HelixSurf is illustrated in \cref{fig:architecture}. 
Notably, SDF-MLP uses Softplus (\ie $\texttt{Softplus}(x) = \frac{1}{\beta} \times \log(1 + \exp(\beta \times x))$) as activation functions, where $\beta = 100$.
Specifically, we apply positional encoding $\gamma(\cdot)$ \cite{mildenhall2020nerf} to the input spatial position $\bm{x}$ as \cref{eq:pos_enc} and apply spherical encoding $\text{Sh}(\cdot)$ \cite{yu2021plenoctrees} to the input view direction $\bm{v}$.
\begin{equation}
    \label{eq:pos_enc}
    \resizebox{.9\hsize}{!}{$\gamma(\bm{x}) = (\sin(2^0\pi \bm{x}), \cos(2^0\pi \bm{x}), \cdots, \sin(2^{L-1}\pi \bm{x}), \cos(2^{L-1}\pi \bm{x}))$}
\end{equation}
We set the frequencies $L$ in positional encoding to 6 and set the degrees of spherical encoding to 4.

\vspace{-0.4cm}

\vspace{-0.4cm}

\section{Handling of Textureless Surface Areas}
\begin{figure*}
    \centering
    \includegraphics[width=0.9\textwidth]{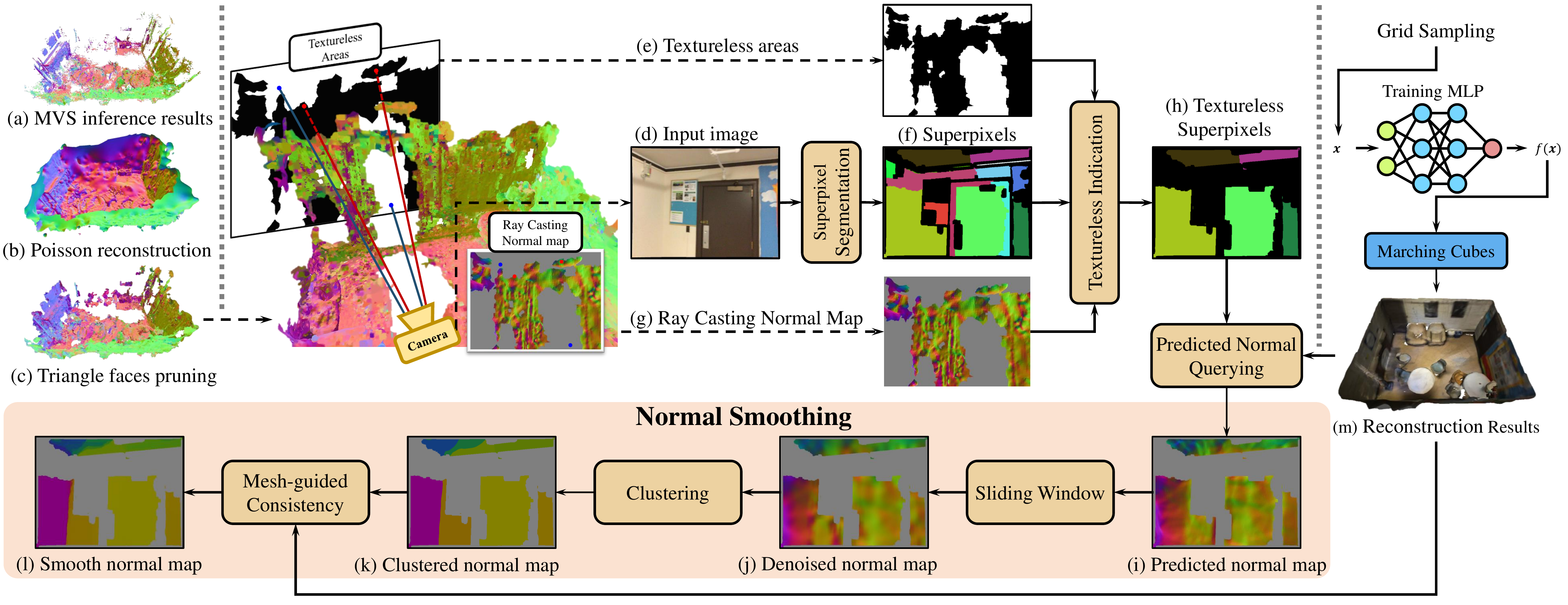}
    \vspace{-0.3cm}
    \caption{\textbf{The overall textureless surface areas handling scheme.}}
\label{fig:normal_smoothing}
\end{figure*}

In this work, we treat the areas that the MVS predictions (\cref{fig:normal_smoothing}(a-c)) are less reliable as textureless surface areas (marked as white regions in \cref{fig:normal_smoothing}(e)), and leverage the homogeneity inside individual superpixels to handle these areas. 
In this scheme, we assume that the superpixels can geometrically partition textureless surface areas. In fact, superpixels (\cref{fig:normal_smoothing}(f)) not only fall in textureless areas but also fall in texture-rich areas or are partially covered by both areas. We thus treat the superpixels (\spa \ \spb \ \spc \ \spd \ in \cref{fig:normal_smoothing}(h)) that are mainly covered by textureless surface areas as the textureless superpixels. Moreover, we treat the superpixel that (\cref{fig:normal_smoothing}(h)) whose smoothness score (\cf. \cref{alg:smot_score})
over 0.9 as a textureless superpixel (\spe \ in \cref{fig:normal_smoothing}(h)) to strengthen the regularization.
After the identification of textureless surface areas by superpixels, we correspondingly querying the predicted normal map (\cref{fig:normal_smoothing}(i)) from the learned MLP of HelixSurf and denoise the predicted normal with a sliding window manner (\cref{fig:normal_smoothing}(j)). \footnote{For those textureless surface areas totally not covered by the MVS predictions, we initialize them with normals generated with Manhattan assumption \cite{guo2022neural}.} 

The primary problem is that the photometric homogeneity inside individual superpixels may not support the correct partition of textureless surface areas at the geometric level (\eg, \spb \ in \cref{fig:normal_smoothing}(h) confuses the corners).
Based on the observation that such superpixels have low smoothness scores (\cf. \cref{alg:smot_score}), we thus conduct the adaptive K-means clustering algorithm (\cf \cref{alg:ada_k_means}) on all textureless superpixels, which adaptively extracts the principal normals for the superpixels. Then, we assign the internal pixels in each textureless superpixels with their corresponding principal normals and obtain the clustered normal map (\cref{fig:normal_smoothing}(k)).
We further consider the consistency among multi-view images and conduct mesh-guided consistency on clustered normal maps.
Finally, the smooth normal maps (\cref{fig:normal_smoothing}(l)) are used to regularize the learning of the neural implicit surface learning in HelixSurf.
The overall textureless surface areas handling scheme is illustrated in \cref{fig:normal_smoothing}.

\begin{algorithm}[htb]
    \caption{Pseudo code of smoothness scores $\delta(\cdot)$}
    \label{alg:smot_score}
    \begin{algorithmic}[1]
    \renewcommand{\algorithmicrequire}{\textbf{Input: }}
    \Require normals $\{\bm{n}\}$ for one superpixel $\mathcal{A}$
    \renewcommand{\algorithmicrequire}{\textbf{Initialize: }}
    \Require{$\text{count}=0$, output$=0$, $\bm{n}_\text{\tiny{mean}} = \frac{1}{\vert\{\bm{n}\}\vert}\sum{\bm{n}}$}
    \For{$\bm{n}$ in $\{\bm{n}\}$}
        \If{$\frac{\bm{n}\cdot\bm{n}_\text{\tiny{mean}}}{\vert\bm{n}\vert\vert\bm{n}_\text{\tiny{mean}}\vert} > 0.9$}
            \State $\text{count} = \text{count} + 1$
        \EndIf
    \EndFor
    \State output = count / $\vert\{\bm{n}\}\vert$
    \renewcommand{\algorithmicrequire}{\textbf{Return: }}
    \Require{output}
    \end{algorithmic}
\end{algorithm}

\begin{algorithm}[htb]
    \caption{Pseudo code of adaptive K-means clustering}
    \label{alg:ada_k_means}
    \begin{algorithmic}[1]
        \renewcommand{\algorithmicrequire}{\textbf{Input: }}
        \Require normals $\{\bm{n}\}$ for one superpixel $\mathcal{A}$, \par
                 smoothness threshold $\tau_{\bm{n}}=0.9$, \par
                 maximum clustering $k_\text{\tiny{max}}=3$ of \texttt{K-means}
        \renewcommand{\algorithmicrequire}{\textbf{Initialize: }}
        \Require{$k=1$, output=$\emptyset$}
        \While{$k\leq k_\text{\tiny{max}}$}
            \State $\{\{\bm{n}\}_j\}_{j=1}^k = \texttt{K-means}(\{\bm{n}\}, k)$
            \If{$\forall\{\delta(\{\bm{n}\}_j)> \tau_{\bm{n}}\}_{j=1}^k$}
                \For{$\{\bm{n}_j\}$ in $\{\{\bm{n}\}_j\}_{j=1}^k$}
                    \State $\bm{n}_\text{\tiny{principal}} = \frac{1}{|\{\bm{n}_j\}|} \sum \bm{n}_j$
                    \State push $\bm{n}_\text{\tiny{principal}}$ into output
                \EndFor
                \State break
            \Else
                \State $k=k+1$
            \EndIf
        \EndWhile
        \renewcommand{\algorithmicrequire}{\textbf{Return: }}
        \Require{output}
    \end{algorithmic}
\end{algorithm}

\begin{figure}[htbp]
    \centering
    \includegraphics[width=0.4\textwidth]{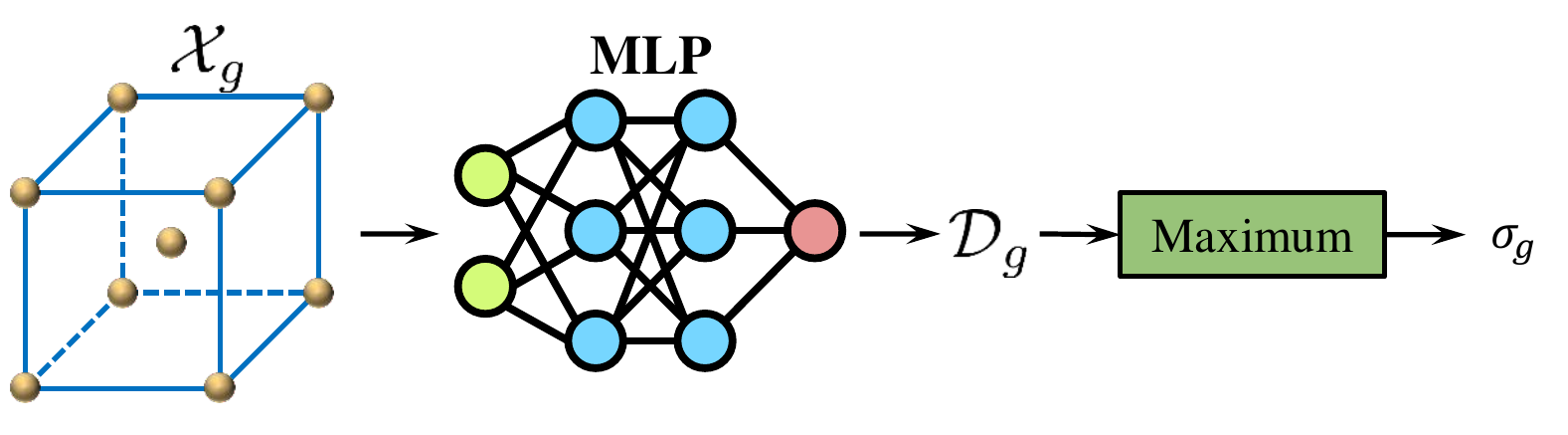}
    \vspace{-0.3cm}
    \caption{Illustration of calculating the density $\sigma_g$ of grid $g$. Note that we follow \cite{wang2021neus} to model density $\sigma$ as an SDF-induced volume density.}
    \label{fig:occupancy_grid}
\end{figure}

\vspace{-0.4cm}

\section{Improving the Efficiency by Establishing Dynamic Space Occupancies}
In this work, we devise a scheme that can adaptively guide the point sampling along rays by maintaining dynamic occupancy grids $\mathcal{G}_\text{\tiny{Occu}}$ in the 3D scene space.

Given the grid $g\subset \mathcal{G}_\text{\tiny{Occu}}$, we update the occupancy $o_g$ of grid $g$ using exponential moving average (EMA), \ie, $o_g^{\text{\tiny EMA}} \leftarrow \max(\sigma_g, \alpha (\sigma_g - o_g^{\text{\tiny EMA}}) + o_g^{\text{\tiny EMA}} )$, where $\sigma_g$ is the density at $g$ by the inducing SDF function $f$ and $\alpha = 0.05$.
To calculate the density of $g$, we set a point set $\mathcal{X}_g = \{\bm{x}_i^g \in \mathbb{R}^3\}_{i=1}^9$ that contains the center and 8 vertices of this grid. Then, we get the predicted densities $\mathcal{D}_g = \{\sigma_i^g \in \mathbb{R}^+\}_{i=1}^9$ of these points and take the maximum of $\mathcal{D}_g$ as the density of grid $\sigma_g$, as illustrated in \cref{fig:occupancy_grid}.

\begin{figure*}
    \centering
    \includegraphics[width=0.9\textwidth]{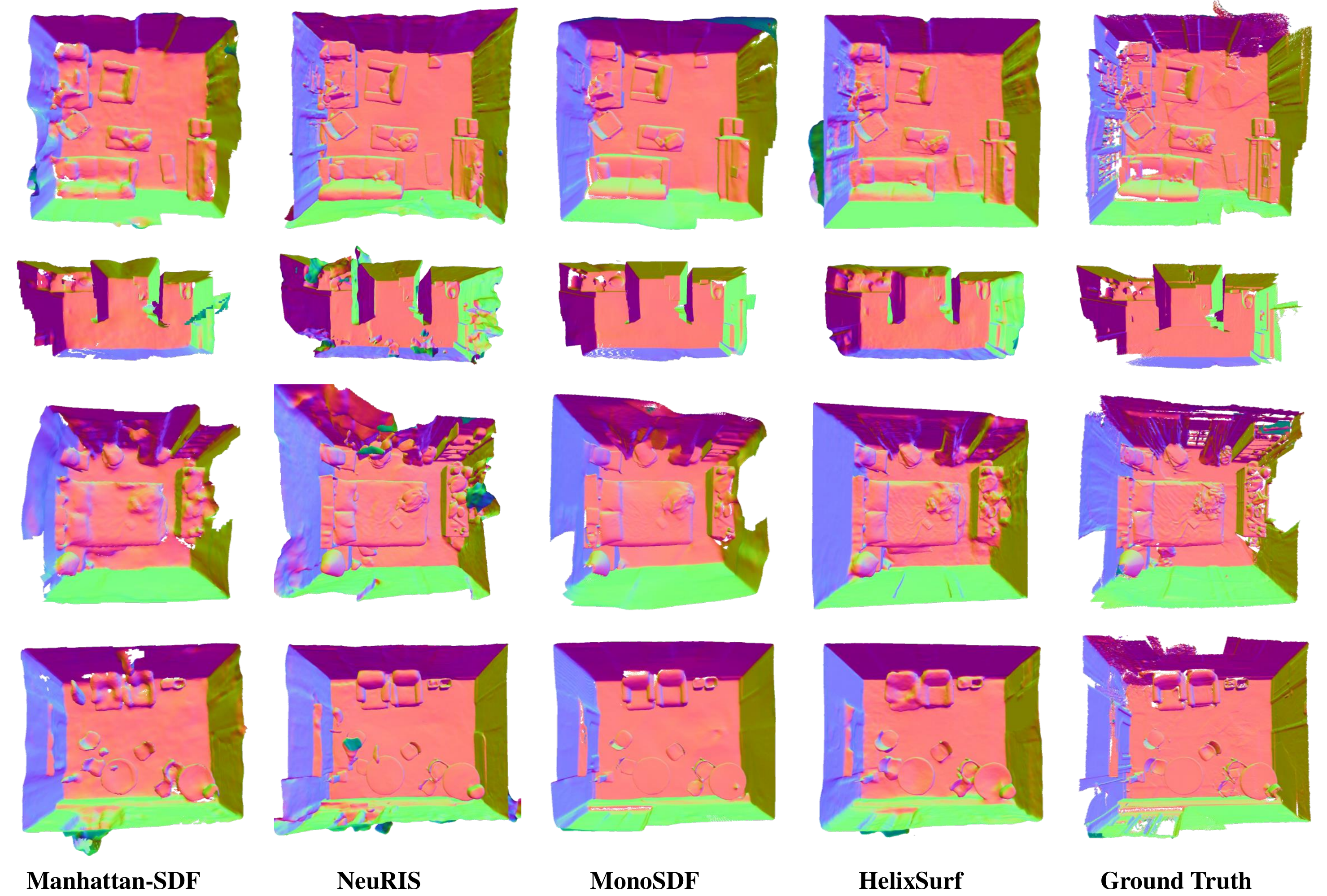}
    \vspace{-0.3cm}
    \caption{\textbf{The top views of the reconstructed scenes on ScanNet.}}
\label{fig:scannet_addition1}
\end{figure*}

\vspace{-2mm}
\section{More Implementation Details}
\label{sec:more_imp_detail}
In this section, we provide more implementation details about the PatchMatch based multi-view stereo (PM-MVS) method, ray casting technique, textureless triangle faces pruning, and the experimental settings.

\vspace{-2mm}
\subsection{Multi-View Stereo}
\label{subsec:multi_view_stereo}
PM-MVS methods \cite{schonberger2016pixelwise, zheng2014patchmatch, galliani2015massively, xu2018multi, xu2020planar} consist of three parts: initialization, iterative sampling \& propagation and fusion.
The stereo fusion algorithm in COLMAP \cite{schonberger2016pixelwise} is time-consuming due to its inefficient interleaved row-/column-wise propagation.
To this end, we integrate the ACMH \cite{xu2018multi} in the HelixSurf. 
ACMH is also the basis MVS scheme in ACMP \cite{xu2020planar}.
On the one hand, AMCH is based on a checkerboard propagation pattern, which achieves higher parallelization. On the other hand, its proposed adaptively sampling scheme makes the entire mechanism efficient and reliable.
While promising, the ordinary ACMH implementation spends a lot of time on I/O operations, and the final fusion step is serially implemented. In this work, we redesign the I/O operations and implement the final fusion step via CUDA kernels.

\subsection{Ray Casting}
\label{subsec:ray_casting}
In this work, we apply ray casting to query normal maps on the reconstructed mesh.
Intuitively, it's a resource-consuming process since hundreds of thousands of rays need to be cast for each map size of $640 \times 480$.
For efficiency, we use NVIDIA OptiX \cite{parker2010optix} technique, a general-purpose ray tracing engine that combines a programmable ray tracing pipeline with a lightweight scene representation.
This technique enables us to customize a parallel ray casting program and render hundreds of normal maps in just seconds with an NVIDIA RTX 3090 GPU.

\subsection{Textureless Triangle Faces Pruning}
\label{subsec:piece_prune}
For handling textureless surface areas, we utilize the inference results of the integrated PM-MVS method to identify textureless surface areas.
Even MVS methods can apply some continuous fitting method (\eg, Poisson reconstruction \cite{kazhdan2006poisson, kazhdan2013screened}) to recover a complete surface (\ie, a set of triangle faces) from their inference results (\ie, discrete points), they fail to recover the correct surface of textureless areas due to the lack of inference results on these areas.
As investigated in \cite{huang2022surface}, the reconstructed surfaces produce a convex hull or concave envelope results in points missing regions and reconstruct isolated components for noises and outliers.
We thus calculate the distance of each triangle face to the nearest point from the inference results, and prune the triangle faces away from the inference results.
Then, we remove the isolated components whose diameter is smaller than a specified constant.
After pruning the textureless triangle faces, the pruned mesh is used to handle the textureless areas.

\subsection{More Experimental Settings}
\label{subsec:exp_setting}
For each scene of ScanNet \cite{dai2017scannet}, we uniformly sample one-tenth of views from the frames of the corresponding video, obtain about 200$\sim$500 images and resize them to the size of $640\times480$ resolution.
For Tanks and Temples \cite{knapitsch2017tanks}, we use all the images from the provided images set, and resize the images to the size of $960\times540$ resolution.
For both datasets, we follow MVSNet \cite{yao2018mvsnet} to choose the neighbor referencing images for each view.

\section{Evaluation Metrics}
\label{sec:evaluation}
In this work, we use the following metrics to evaluate the reconstruction quality: \textit{Accuracy}, \textit{Completeness}, \textit{Precision}, \textit{Recall}, and \textit{F-score}. The definitions of these metrics are shown in \cref{table:metrics}.
And the metrics for evaluating depth and normal map are shown in \cref{table:metrics_depthnorm}.

\begin{table}[!htb]
    \centering
    \scalebox{1.0}{
    \begin{tabular}{lc}
        \Xhline{2pt} \\
        Metric & Definition \\
        \Xhline{1pt} \\
        Accuracy &
        $ \text{mean}_{p\in P} (\min_{p^*\in P^*} \Vert p - p^* \Vert)$ \\[15pt]
        Completeness &
        $ \text{mean}_{p^*\in P^*} (\min_{p\in P} \Vert p - p^* \Vert)$ \\[15pt]
        Precision &
        $ \text{mean}_{p\in P} (\min_{p^*\in P^*} \Vert p - p^* \Vert < 0.05)$ \\[15pt]
        Recall &
        $ \text{mean}_{p^*\in P^*} (\min_{p\in P} \Vert p - p^* \Vert < 0.05)$ \\[15pt]
        F-score &
        $\frac{2 \times \text{Precision} \times \text{Recall}}{\text{Precision} + \text{Recall}}$ \\[15pt]
        \Xhline{2pt}
    \end{tabular}
    }
    \caption{\textbf{Evaluation metrics for reconstruction quality used in this work}. $P$ and $P^*$ are the points sampled from the predicted and the ground truth mesh.\label{table:metrics}}
\end{table}

\begin{table}[!htb]
    \centering
    \scalebox{0.63}{
    \begin{tabular}{llc}
        \Xhline{2pt} \\
        \multicolumn{2}{l}{\makecell{Metric}} & Definition \\
        \Xhline{1pt} \\
        & Abs Diff & 
        $ \frac{1}{n}\sum{|d - d^*|} $ \\[10pt]
        \multirow{4}{*}{\rotatebox[origin=c]{90}{Depth map}} & Abs Rel &
        $ \frac{1}{n}\sum{\frac{|d - d^*|}{d^*}} $ \\[10pt]
        & Sq Rel &
        $ \frac{1}{n}\sum{\frac{|d - d^*|^2}{d^*}} $ \\[10pt]
        & RMSE &
        $ \sqrt{\frac{1}{n}\sum{|d - d^*|^2}} $ \\[10pt]
        \Xhline{2pt}
    \end{tabular}
    \hspace{-6pt}
    \scalebox{0.976}{
    \begin{tabular}{llc}
        \Xhline{2pt} \\
        \multicolumn{2}{l}{\makecell{Metric}} & Definition \\
        \Xhline{1pt} \\
        & Mean & 
        $ \frac{1}{n}\sum{\cos^{-1}[\frac{|\bm{n}\cdot\bm{n}^*|}{|\bm{n}||\bm{n}^*|}]} $ \\[10pt]
        \multirow{4}{*}{\rotatebox[origin=c]{90}{Normal map}} & Median &
        $ \text{median}\left\{\cos^{-1}[\frac{|\bm{n}\cdot\bm{n}^*|}{|\bm{n}||\bm{n}^*|}]\right\} $ \\[10pt]
        & RMSE &
        $ \sqrt{\frac{1}{n}\sum{(\cos^{-1}[\frac{|\bm{n}\cdot\bm{n}^*|}{|\bm{n}||\bm{n}^*|}])^2}} $ \\[10pt]
        & Prop\_30$^{\circ}$ &
        $ \frac{1}{n}\#\left\{{\bm{n}, \bm{n}^*:\cos^{-1}[\frac{|\bm{n}\cdot\bm{n}^*|}{|\bm{n}||\bm{n}^*|}] < 30^{\circ}}\right\} $ \\[10pt]
        \Xhline{2pt}
    \end{tabular}
    }}
    \caption{\textbf{Evaluation metrics for depth and normal map used in this work.} 
    $n$ is the number of pixels with valid depth or normal in ground truth (GT) depth map or normal map. 
    $d$ and $d^*$ are the predicted and GT depths.
    $\bm{n}$ and $\bm{n}^*$ are the predicted and GT normals.
    }
    \label{table:metrics_depthnorm}
\end{table}

\section{Additional Results}
\label{sec:additional_restuls}
In this section, we provide more experimental results for the ScanNet dataset \cite{dai2017scannet} and Tanks \& Temples \cite{knapitsch2017tanks} dataset.
Further, we conduct HelixSurf for object-level and real-world scene reconstruction.
More visualization details are shown in the attached video.

\subsection{ScanNet}
\label{subsec:scannet}
\begin{figure*}
    \centering
    \includegraphics[width=1.0\textwidth]{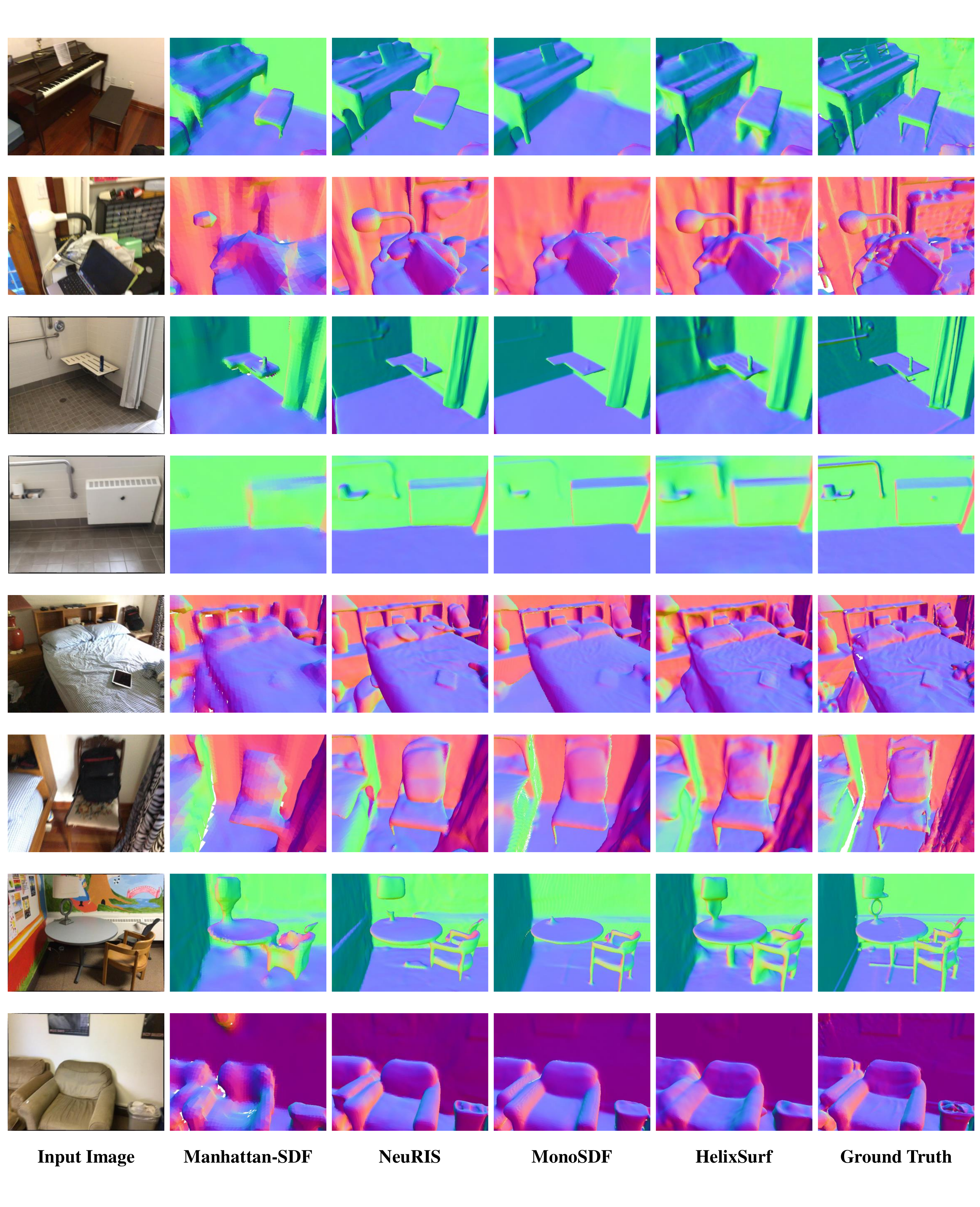}
    \caption{The zoom-in views of the reconstructed scenes on ScanNet.}
\label{fig:scannet_addition2}
\end{figure*}

We show more qualitative results in \cref{fig:scannet_addition1} and \cref{fig:scannet_addition2}. Compared to the state-of-the-art learning-based methods, HelixSurf produces better reconstructions.

\subsection{Tanks and Temples}
\label{subsec:TnT}
\begin{figure*}
    \centering
    \includegraphics[width=1.0\textwidth]{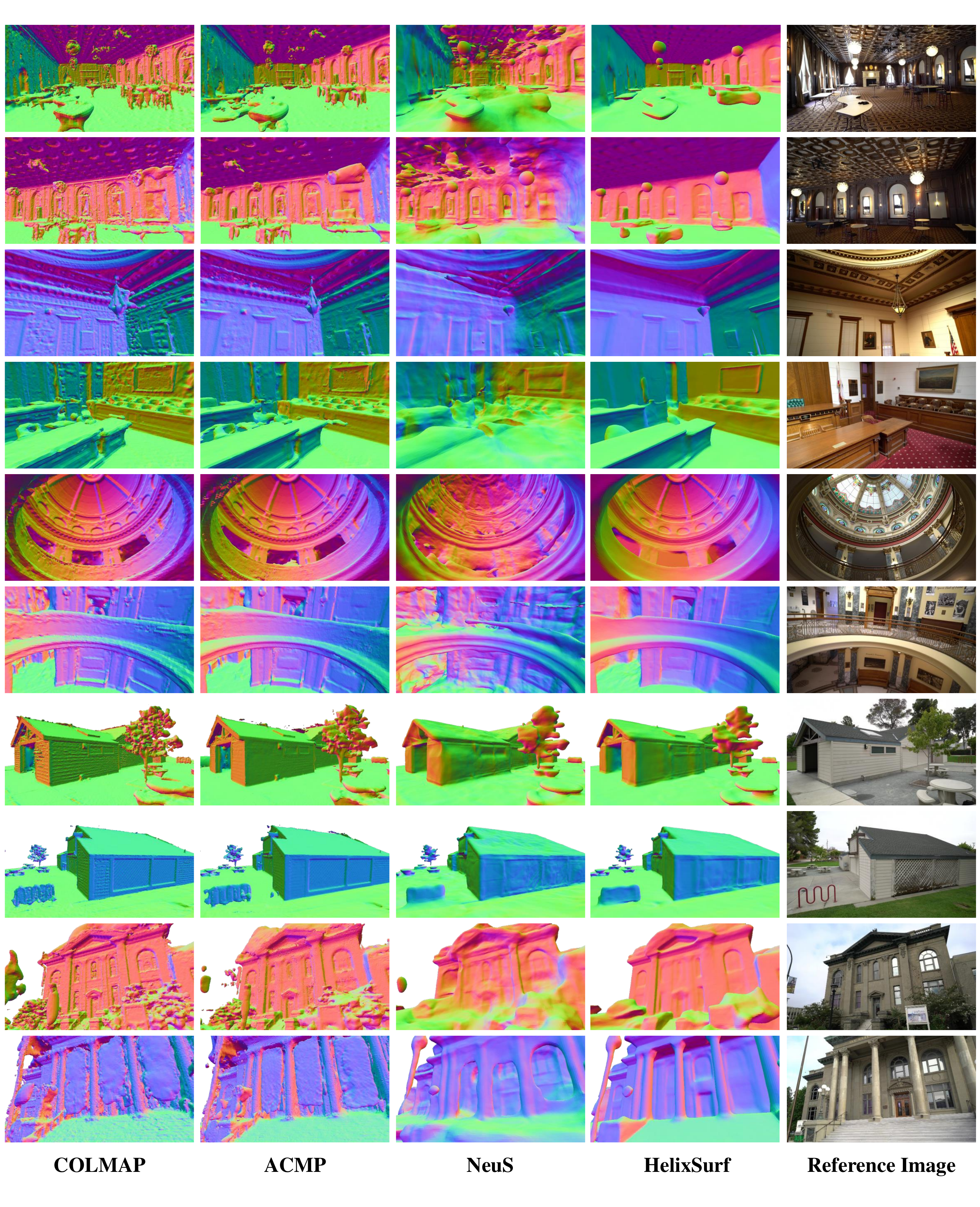}
    \caption{Qualitative Comparison on Tanks and Temples.}
\label{fig:TnT_qua}
\end{figure*}

We show more qualitative results on Tanks and Temples \cite{knapitsch2017tanks} in \cref{fig:TnT_qua}.
Our method can produce more precise and complete geometry than baseline methods.

\subsection{Object-level reconstruction}
\label{subsec:object}
\begin{figure*}
    \centering
    \includegraphics[width=1.0\textwidth]{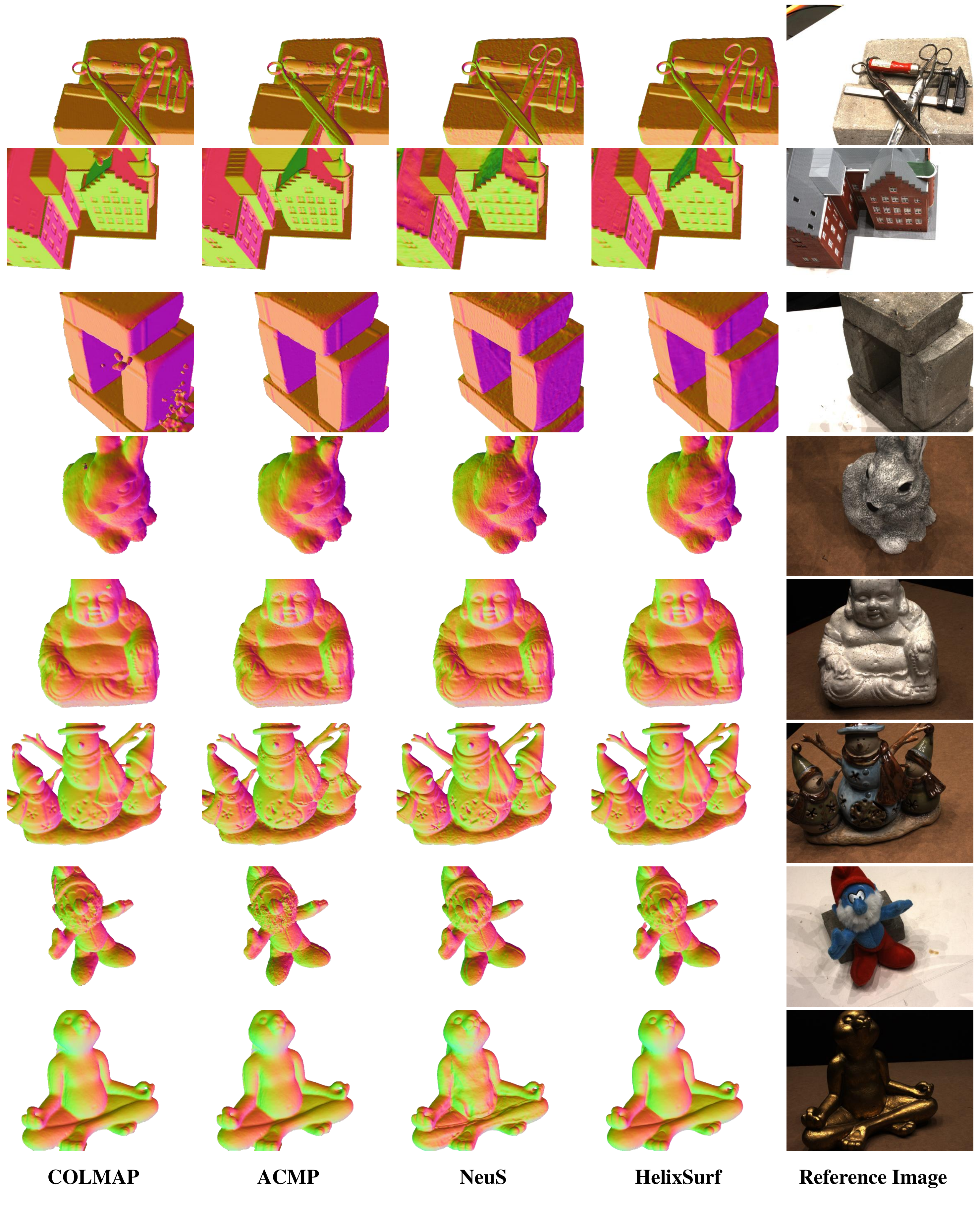}
    \caption{Qualitative Comparison on DTU.}
\label{fig:dtu}
\end{figure*}

Although our HelixSurf is proposed for scene-level reconstruction, we examine its reconstruction on an object-level dataset (\ie DTU \cite{aanaes2016large}) and report the result in \cref{fig:dtu}.
As can be seen that HelixSurf stands up against baselines.

\subsection{Real-World Scene}
\label{subsec:real_world_scene}
\begin{figure*}
    \centering
    \begin{minipage}[t]{0.51\linewidth}
        \centering
        \includegraphics[width=\textwidth]{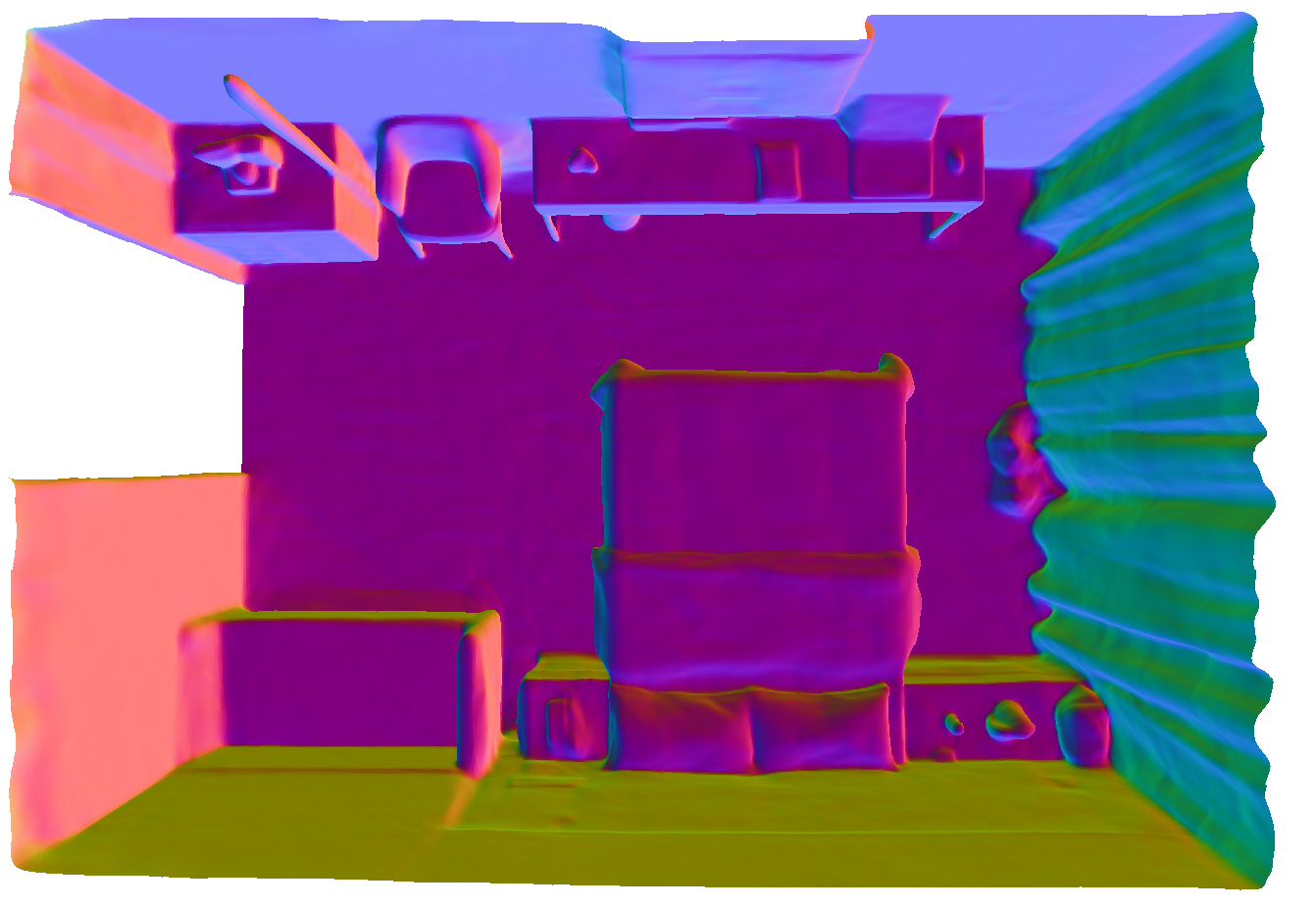}
        \subcaption{The top view of the entire reconstructed indoor scene.}
        \label{fig:real_world1}
    \end{minipage} \quad
    \begin{minipage}[t]{0.45\linewidth}
        \includegraphics[width=\textwidth]{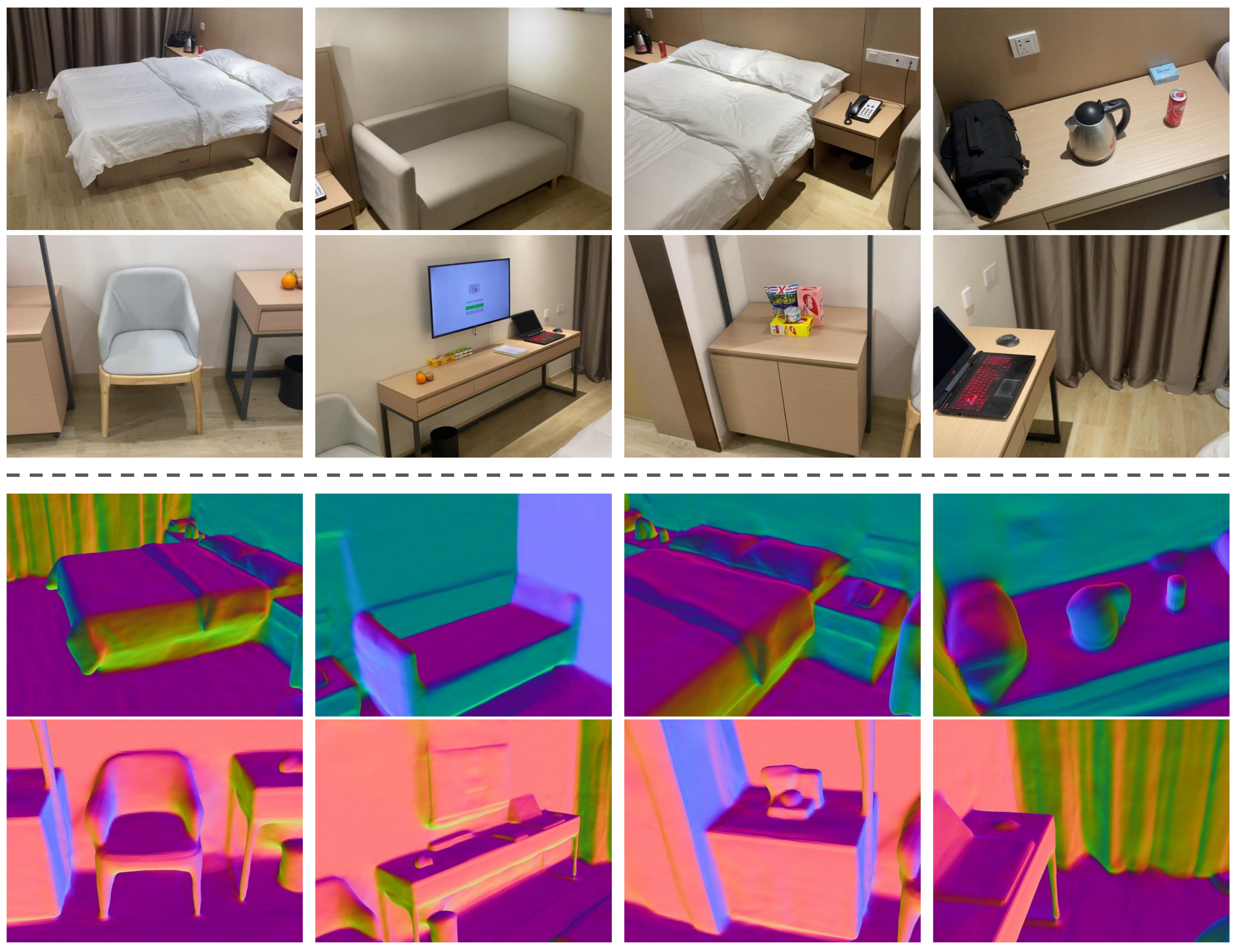}
        \subcaption{The first block is a set of reference images. The second block is the corresponding reconstruction results. Surface normals are visualized as coded colors.}
        \label{fig:real_world2}
    \end{minipage}
    \caption{\textbf{Reconstruction of the real-world capture.}}
\label{fig:real_world}
\end{figure*}

In order to demonstrate the efficacy of HelixSurf in the real-world capture, we conduct HelixSurf on the real-world collected image set (captured by iPhone 11 in \cite{wang2022neuris}). The qualitative reconstruction result is shown in \cref{fig:real_world}.

\section{Novel View Synthesis}
\label{sec:nvs}
\begin{figure*}
    \centering
    \includegraphics[width=0.95\textwidth]{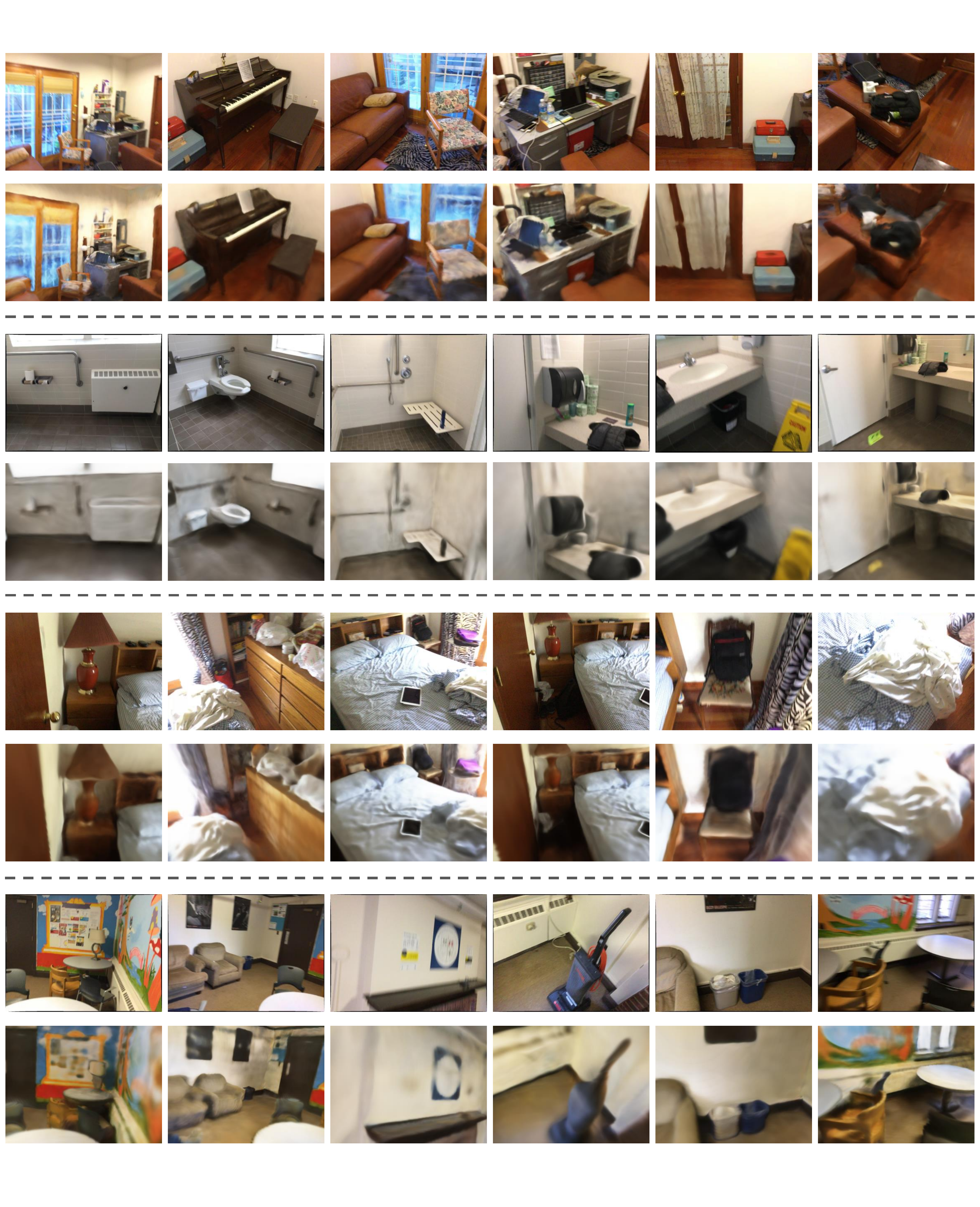}
    \caption{\textbf{Novel view synthesis results on ScanNet.} For each block, the first row is the reference images and the second row is the novel view synthesis results.}
\label{fig:nvs}
\end{figure*}

In this work, we aim to achieve an accurate and complete reconstruction of the target scene. Furthermore, the accurate reconstruction results enable us to realize high-quality novel view synthesis. For reconstruction, we uniformly sample one-tenth of views from the target scene in ScanNet \cite{dai2017scannet}. For the novel view synthesis, we randomly select some views from the remaining nine-tenths of views and conduct rendering. The rendering results are shown in \cref{fig:nvs}.

\section{Failure Cases}
\label{sec:failure}
\begin{figure}[htbp]
    \centering
    \begin{minipage}[t]{0.46\linewidth}
        \centering
        \includegraphics[width=\textwidth]{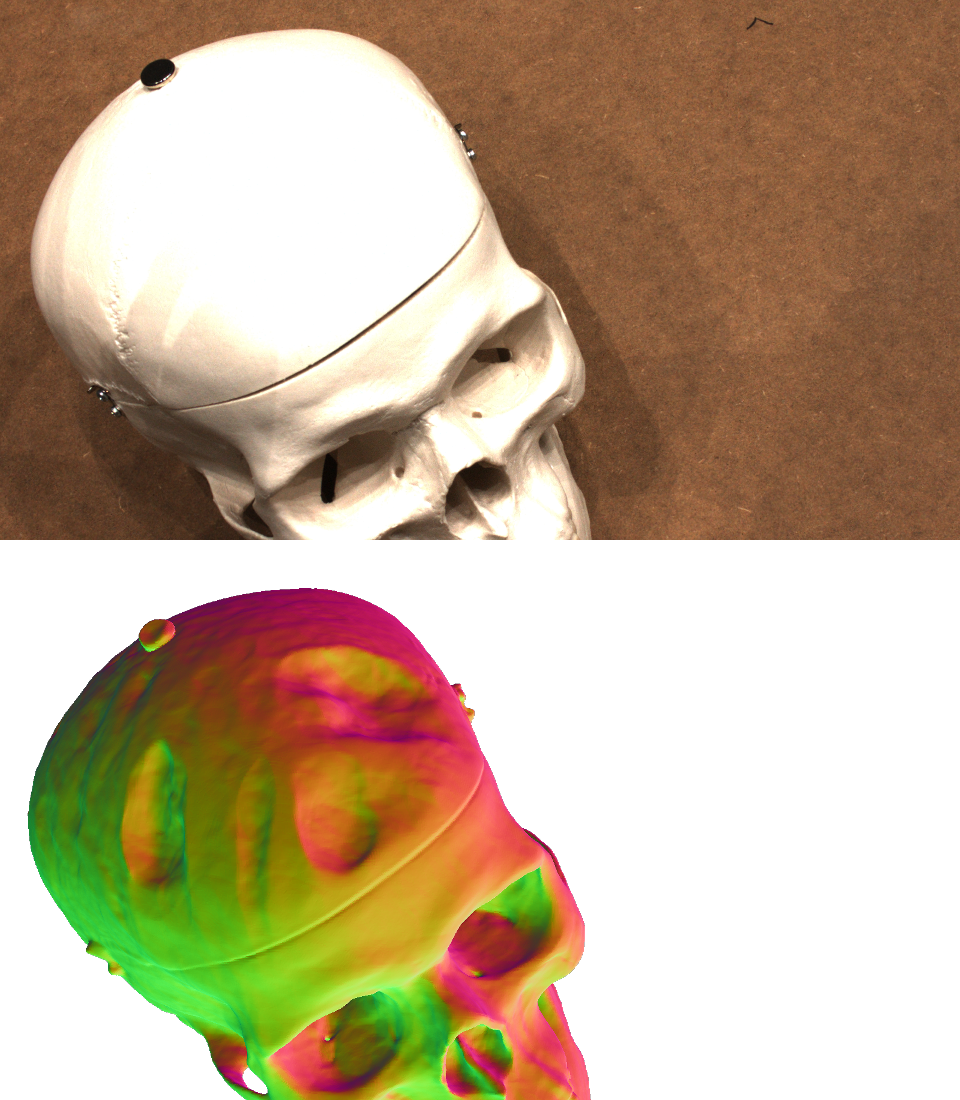}
        \subcaption{}
        \label{fig:failure_case1}
    \end{minipage} \ \
    \begin{minipage}[t]{0.46\linewidth}
        \includegraphics[width=\textwidth]{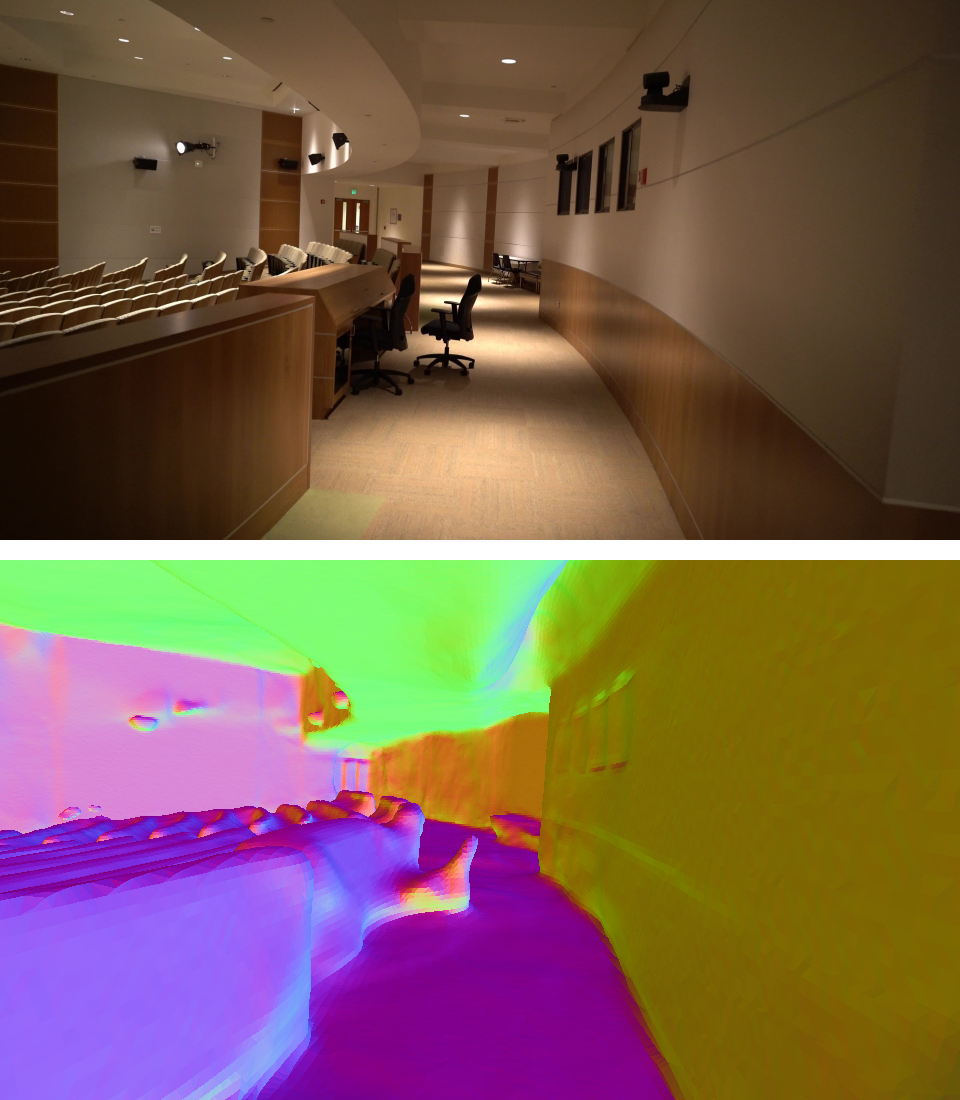}
        \subcaption{}
        \label{fig:failure_case2}
    \end{minipage}
    \caption{\textbf{Failure cases.} the first row is the reference images and the second row is the reconstruction results. Surface normals are visualized as coded colors. (a) and (b) show the failure cases at the object-level and the scene-level, respectively.}
    \label{fig:failure}
\end{figure}

In this work, we assume that textureless surface areas tend to be both homogeneous in color and geometrically smooth. Once the textureless surface areas in the scene do not satisfy this assumption, HelixSurf may fail to handle these areas.
For the textureless surface areas with significant curvature as shown in \cref{fig:failure}, HelixSurf may fail to handle these textureless surface areas by the normal smoothing scheme and suffer from the artifacts in the reconstruction results.
These problems can be solved by adopting some geometric assumptions about the curved surface. However, it will undoubtedly increase the complexity of the whole system. An interesting future work is to introduce more flexible and generalized assumptions to tackle the corner cases.

\newpage

\end{document}